\documentclass[twoside,11pt]{article}

\newcommand{\xv}{\mathbf{x}}

\newcommand{\yv}{\mathbf{y}}

\newcommand{\zv}{\mathbf{z}}

\newcommand{\fv}{\mathbf{f}}

\newcommand{\wv}{\mathbf{w}}
\newcommand{\Wv}{\mathbf{W}}

\newcommand{\ud}{\mathrm{d}}

\usepackage{jmlr2e}
\usepackage{enumerate}
\usepackage{algorithm}
\usepackage{algorithmic}
\usepackage{epsf}
\usepackage{multirow}
\usepackage{wrapfig}
\usepackage{graphicx}
\usepackage{subfigure} 

\jmlrheading{1}{2008}{1-48}{4/00}{10/00}{Jun Zhu, Amr Ahmed, and Eric P. Xing}

\ShortHeadings{MedLDA: A General Framework of Maximum Margin Supervised Topic Models}{Zhu, Ahmed,
and Xing} \firstpageno{1}

\begin{document}

\title{MedLDA: A General Framework of Maximum Margin Supervised Topic Models}

\author{\name Jun Zhu \email junzhu@cs.cmu.edu \\
        \addr School of Computer Science \\
       Carnegie Mellon University\\
       \AND
        \name Amr Ahmed  \email amahmed@cs.cmu.edu\\
       \addr School of Computer Science \\
       Carnegie Mellon University \\
        \AND
       \name Eric P. Xing  \email epxing@cs.cmu.edu\\
       \addr School of Computer Science \\
       Carnegie Mellon University}

\editor{?}

\maketitle

\begin{abstract}

Supervised topic models utilize document's side information for
discovering predictive low dimensional representations of documents.
Existing models apply the likelihood-based estimation. In this
paper, we present a general framework of max-margin supervised topic models
for both continuous and categorical response variables. Our approach, the
maximum entropy discrimination latent Dirichlet allocation (MedLDA),
utilizes the max-margin principle to train supervised topic models
and estimate predictive topic representations that are arguably more
suitable for prediction tasks. The general principle of MedLDA can be
applied to perform joint max-margin learning and maximum likelihood estimation
for arbitrary topic models, directed or undirected, and
supervised or unsupervised, when the supervised side information is available.
We develop efficient variational methods for posterior inference and parameter estimation,
and demonstrate qualitatively and quantitatively the advantages of MedLDA over likelihood-based
topic models on movie review and 20 Newsgroups data sets.

\end{abstract}


\section{Introduction}

Latent topic models such as Latent Dirichlet Allocation (LDA)
\citep{Blei:03} have recently gained much popularity in managing a
large collection of documents by discovering a low dimensional
representation that captures the latent semantic of the collection.
LDA posits that each document is an admixture of latent topics where
the topics are represented as unigram distribution over a given vocabulary. The
document-specific admixture proportion is distributed as a latent
Dirichlet random variable and represents a low dimensional
representation of the document. This low dimensional representation
can be used for tasks like classification and clustering or merely
as a tool to structurally browse the otherwise unstructured
collection.

The traditional LDA \citep{Blei:03} is an unsupervised model, and thus is incapable
of incorporating the useful side information associated with corpora,
which is uncommon. For example, online users usually post their
reviews for products or restaurants with a rating score or pros/cons
rating; webpages can have their category labels; and the images in
the LabelMe~\citep{Russell:08} dataset are organized in different
categories and each image is associated with a set of annotation
tags. Incorporating such supervised side information may guide the
topic models towards discovering secondary or non-dominant
statistical patterns \citep{Chechik:02}, which may be more
interesting or relevant to the users' goals (e.g., predicting on
unlabeled data). In contrast, the unsupervised LDA ignores such
supervised information and may yields more prominent and perhaps
orthogonal (to the users' goals) latent semantic structures. This
problem is serious when dealing with complex data, which usually
have multiple, alternative, and conflicting underlying structures.
Therefore, in order to better extract the relevant or interesting
underlying structures of corpora, the supervised side information
should be incorporated.

Recently, learning latent topic models with side information has
gained increasing attention. Major instances include the supervised
topic models (sLDA) \citep{Blei:07} for regression\footnote{Although
integrating sLDA with a generalized linear model was discussed in
\citep{Blei:07}, no result was reported about the performance of
sLDA when used for classification tasks. The classification model
was reported in a later paper \citep{Wang:09}}, multi-class LDA (an
sLDA classification model)~\citep{Wang:09}, and the discriminative
LDA (DiscLDA) \citep{Simon:08} classification model. All these
models focus on the document-level supervised information, such as
document categories or review rating scores. Other variants of
supervised topic models have been designed to deal with different
application problems, such as the aspect rating model
\citep{Titov:08} and the credit attribution model \citep{Ramage:09},
of which the former predicts ratings for each aspect and the latter
associate each word with a label. In this paper, without loss of
generality, we focus on incorporating document-level supervision
information. Our learning principle can be generalized to arbitrary
topic models. For the document level models, although sLDA and
DiscLDA share the same goal (uncovering the latent structure in a
document collection while retaining predictive power for supervised
tasks), they differ in their training procedures. sLDA is trained by
maximizing the joint likelihood of data and response variables while
DiscLDA is trained to maximize the conditional likelihood of
response variables. Furthermore, to the best of our knowledge,
almost all existing supervised topic models are trained by
maximizing the data likelihood.

In this paper, we propose a general principle for learning
\emph{max-margin discriminative} supervised latent topic models for
both regression and classification. In contrast to the two-stage
procedure of using topic models for prediction tasks (i.e., first
discovering latent topics and then feeding them to downstream
prediction models), the proposed {\it maximum entropy discrimination
latent Dirichlet allocation} (MedLDA) is an integration of
max-margin prediction models (e.g., support vector machines for classification)
and hierarchical Bayesian topic models by optimizing a single objective function with a set of {\it expected}
margin constraints. MedLDA is a special instance of PoMEN (i.e.,
partially observed maximum entropy discrimination Markov network)
\citep{Zhu:08b}, which was proposed to combine max-margin learning
and structured hidden variables in undirected Markov networks, for
discovering latent topic presentations of documents. In MedLDA, the
parameters for the regression or classification model are learned in
a max-margin sense; and the discovery of latent topics is coupled
with the max-margin estimation of the model parameters. This
interplay yields latent topic representations that are more
discriminative and more suitable for supervised prediction tasks.

The principle of MedLDA to do joint max-margin learning and maximum likelihood estimation
is extremely general and can be applied to arbitrary topic models,
including directed topic models (e.g., LDA and sLDA) or undirected Markov networks (e.g., the Harmonium
\citep{Welling:nips04}), unsupervised (e.g., LDA and Harmonium) or
supervised (e.g., sLDA and hierarchical Harmonium \citep{Yang:07}), and
other variants of topic models with different priors, such as correlated topic models (CTMs) \cite{Blei:nips05}.
In this paper, we present several examples of applying the max-margin principle to
learn MedLDA models which use the unsupervised and supervised LDA as the underlying topic models to discover latent topic representations of documents for both regression and classification. We develop
efficient and easy-to-implement variational methods for MedLDA, and
in fact its running time is comparable to that of an unsupervised LDA for classification.
This property stems from the fact that the MedLDA classification model directly optimizes the margin and does
not suffer from a normalization factor which generally makes learning hard as in fully generative models such as sLDA.

The paper is structured as follows. Section 2 introduces the basic concepts of latent topic models. Section 3 and Section 4 present the MedLDA models for regression and classification respectively, with efficient variational EM algorithms. Section 5 discusses the generalization of MedLDA to other latent variable topic models. Section 6 presents
empirical comparison between MedLDA and likelihood-based topic
models for both regression and classification. Section 7 presents some related works. Finally, Section 8 concludes this paper with future research directions.

\section{Unsupervised and Supervised Topic Models}
In this section, we review the basic concepts of unsupervised and supervised topic models
and two variational upper bounds which will be used later.

\begin{figure}
\centering
\includegraphics[height=100pt]{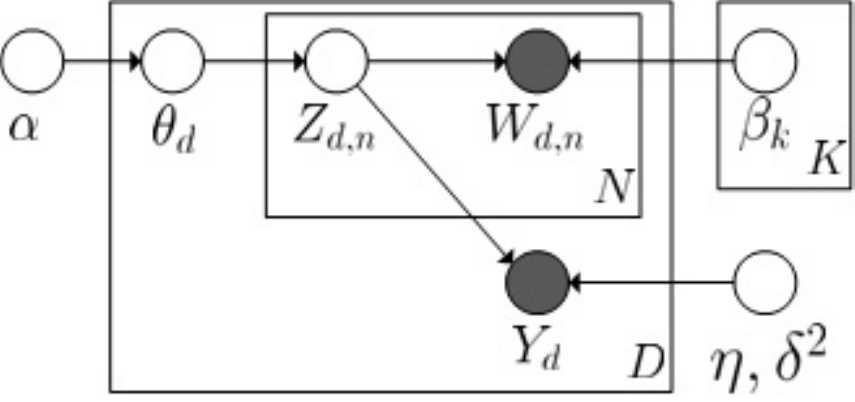}
\caption{Supervised topic model~\citep{Blei:07}.} \label{fig:blei:07}
\end{figure}

The unsupervised LDA (latent Dirichlet allocation) \citep{Blei:03}
is a hierarchical Bayesian model, where topic proportions for a
document are drawn from a Dirichlet distribution and words in the
document are repeatedly sampled from a topic which itself is drawn
from those topic proportions. Supervised topic models (sLDA)
\citep{Blei:07} introduce a response variable to LDA
for each document, as illustrated in Figure~\ref{fig:blei:07}. 

Let $K$ be the number of topics and $M$ be the number of terms in a
vocabulary. $\beta$ denotes a $K \times M$ matrix and each $\beta_k$
is a distribution over the $M$ terms. For the regression problem,
where the response variable $y \in \mathbb{R}$, the
generative process of sLDA is as follows: \\[-0.5cm]

\begin{enumerate}
\item Draw topic proportions $\theta | \alpha \sim \textrm{Dir}(\alpha)$.
\item For each word
    \begin{enumerate}[(a)]
    \item Draw a topic assignment $z_n | \theta \sim
    \textrm{Mult}(\theta)$.
    \item Draw a word $w_n | z_n, \beta \sim
    \textrm{Multi}(\beta_{z_n})$.
    \end{enumerate}
\item Draw a response variable: $y | z_{1:N}, \eta, \delta^2 \sim
N(\eta^\top \bar{z}, \delta^2)$, where $\bar{z} = 1/N \sum_{n=1}^N z_n$ is the average topic proportion of a document.
\end{enumerate}

The model defines a joint distribution:
\begin{displaymath}p(\theta, \zv, \yv, \Wv | \alpha, \beta, \eta, \delta^2) =
\prod_{d=1}^D p(\theta_d | \alpha) ( \prod_{n=1}^N p(z_{dn} |
\theta_d) p(w_{dn} | z_{dn}, \beta) ) p(y_d | \eta^\top \bar{z}_d,
\delta^2),
\end{displaymath}
where $\yv$ is the vector of response variables in a corpus $\mathcal{D}$ and $\Wv$ are all the words.
The joint likelihood on $\mathcal{D}$ is $p(\yv, \Wv
| \alpha, \beta, \eta, \delta^2)$. To estimate the unknown parameters $(\alpha,\beta,\eta,\delta^2)$,
sLDA maximizes the log-likelihood $\log p(\yv, \Wv|\alpha, \beta, \eta,
\delta^2)$. Given a new document, the expected response value is the prediction:
\begin{eqnarray} \hat{y} \triangleq E \lbrack Y | w_{1:N}, \alpha, \beta, \eta,
\delta^2 \rbrack = \eta^\top E\lbrack \bar{Z} | w_{1:N}, \alpha,
\beta, \delta^2 \rbrack, \label{sLDA}
\end{eqnarray}
\noindent where $E\lbrack X \rbrack$ is an expectation with respect to the
posterior distribution of the random variable $X$.

Since exact inference of the posterior distribution of hidden variables and the likelihood is intractable, variational methods~\citep{Jordan:99} are applied to get approximate solutions. Let $q(\theta, \zv | \gamma, \phi)$ be a variational distribution that approximates the posterior $p(\theta, \zv | \alpha, \beta, \eta, \delta^2, \yv, \Wv)$. By using Jensen's inequality, we can get a variational upper bound of the negative log-likelihood:
\begin{displaymath}
\mathcal{L}^s(q) = - E_q\lbrack \log p(\theta, \zv, \yv, \Wv | \alpha, \beta, \eta, \delta^2) \rbrack - \mathcal{H}(q(\zv,
\theta )) \geq - \log p(\yv, \Wv | \alpha, \beta, \eta, \delta^2),
\end{displaymath}
where $\mathcal{H}(q) \triangleq - E_q \lbrack \log q \rbrack$ is the entropy of $q$. By introducing some independence assumptions (like mean field) about the $q$ distribution, this upper bound can be efficiently optimized, and we can estimate the parameters $(\alpha,\beta,\eta,\delta^2)$ and get the best approximation $q$. See \citep{Blei:07} for more details.

For the unsupervised LDA, the generative procedure is similar, but without the third step. The joint distribution is $p(\theta, \zv, \Wv | \alpha, \beta) = \prod_{d=1}^D p(\theta_d | \alpha) ( \prod_{n=1}^N p(z_{dn} | \theta_d) p(w_{dn} | z_{dn}, \beta) )$ and the likelihood is $p(\Wv | \alpha, \beta)$. Similarly, a variational upper bound can be derived for approximate inference:
\begin{displaymath}
\mathcal{L}^u(q) = - E_q\lbrack \log p(\theta, \zv, \Wv | \alpha, \beta) \rbrack - \mathcal{H}(q(\zv, \theta)) \geq - \log p(\Wv | \alpha, \beta ),
\end{displaymath}
where $q(\theta, \zv)$ is a variational distribution that approximates the posterior $p(\theta, \zv | \alpha, \beta, \Wv)$. Again, by making some independence assumptions, parameter estimation and posterior inference can be efficiently done by optimizing $\mathcal{L}^u(q)$. See \citep{Blei:03} for more details.

In sLDA, by changing the distribution model of generating response variables,
other types of responses can be modeled, such as the discrete classification problem \citep{Blei:07,Wang:09}. However, the posterior inference and parameter estimation in supervised LDA classification model are much more difficult than those of the sLDA regression model because of the normalization factor of the non-Gaussian distribution model for response variables. Variational methods or multi-delta methods were used to approximate the normalization factor \citep{Wang:09,Blei:07}. DiscLDA \citep{Simon:08} is a discriminative variant of supervised topic models for classification, where the unknown parameters (i.e., a linear transformation matrix) are learned by maximizing the conditional likelihood of the response variables.

Although both maximum likelihood estimation (MLE) and maximum conditional likelihood estimation (MCLE) have shown great success in many cases, the max-margin learning is arguably more discriminative and closer to our final prediction task in supervised topic models. Empirically, max-margin methods like the support vector machines (SVMs) for classification have demonstrated impressive success in a wide range of tasks, including image classification, character recognition, etc. In addition to the empirical success, max-margin methods enjoy strong generalization guarantees, and are able to use kernels, allowing
the classifier to deal with a very high-dimensional feature space.

To integrate the advantages of max-margin methods into the procedure of discovering latent topics, below, we present a max-margin variant of the supervised topic models, which can discover predictive topic representations that are more suitable for supervised prediction tasks, e.g., regression and classification.

\section{Maximum Entropy Discrimination LDA for Regression}

In this section, we consider the supervised prediction task, where the response variables take continuous real values. This is known as a regression problem in machine learning. We present two MedLDA regression models that perform max-margin learning for the supervised LDA and unsupervised LDA models. Before diving into the full exposition of our methods, we first review the basic support vector regression method, upon which MedLDA is built.

\subsection{Support Vector Regression}

Support vector machines have been developed for both classification and regression.
In this section, we consider the support vector regression (SVR), on which a comprehensive tutorial has been published by \cite{Smola:03}. Here, we provide a brief recap of the basic concepts.

Suppose we are given a training set $\mathcal{D}=\{ (\xv_1, y_1), \cdots, (\xv_D, y_D) \}$, where $\xv \in \mathcal{X}$ are inputs and $y \in \mathbb{R}$ are real response values. In $\epsilon$-support vector regression \citep{Vapnik:95}, our goal is to find a function $h(\xv) \in \mathcal{F}$ that has at most $\epsilon$ deviation from the true response values $y$ for all the training data, and at the same time as flat as possible. One common choice of the function family $\mathcal{F}$ is the linear functions, that is, $h(\xv) = \eta^\top \fv(\xv)$, where $\fv = \{ f_1, \cdots, f_K \}$ is a vector of feature functions. Each $f_k: \mathcal{X} \to \mathbb{R}$ is a feature function. $\eta$ is the corresponding weight vector. Formally, the linear SVR finds an optimal linear function by solving the following constrained convex optimization problem
\setlength\arraycolsep{2pt}\begin{eqnarray} {\rm
P0(SVR)}: & & \min_{\eta, \xi, \xi^\star}~ \frac{1}{2} \Vert \eta \Vert_2^2 + C \sum_{d=1}^D (\xi_d + \xi_d^\star) \nonumber \\
\mathrm{s.t.}~\forall d: & & \left\{ \begin{array}{rl}
    y_d - \eta^\top \fv(\xv_d) & \leq \epsilon + \xi_d \\
    - y_d + \eta^\top \fv(\xv_d) & \leq \epsilon + \xi_d^\star \\
    \xi_d, \xi_d^\star & \geq 0
    \end{array} \right. ,\nonumber
\end{eqnarray}
\noindent where $\Vert \eta \Vert_2^2 = \eta^\top \eta$ is the $\ell_2$-norm; $\xi$ and $\xi^\star$ are slack variables that tolerates some errors in the training data; and $\epsilon$ is the precision parameter. The positive regularization constant $C$ determines the trade-off between the flatness of $h$ (represented by the $\ell_2$-norm) and the amount up to which deviations larger than $\epsilon$ are tolerated. The problem P0 can be equivalently formulated as a regularized empirical loss minimization, where the loss is the so-called $\epsilon$-insensitive loss \citep{Smola:03}.

For the standard SVR optimization problem, P0 is a QP problem and can be easily solved in the dual formulation.
In the Lagrangian method, samples with non-zero lagrange multipliers are called support vectors, the same as in SVM classification model. There are also some freely available packages for solving a standard SVR problem, such as the SVM-light \citep{Joachims:99}. We will use these methods as a sub-routine to solve our proposed approach.

\subsection{Learning MedLDA for Regression}

Instead of learning a point estimate of $\eta$ as in sLDA, we take a more general \footnote{Under the special case of linear models, the posterior mean of an averaging model can be directly solved in the same manner of point estimate.} Bayesian-style (i.e., an averaging model) approach and learn a distribution\footnote{In principle, we can perform Bayesian-style estimation for other parameters, like $\delta^2$. For simplicity, we only consider $\eta$ as a random variable in this paper.} $q(\eta)$ in a max-margin manner. For prediction, we take the average over all the possible models (represented by $\eta$):
\begin{eqnarray}
\hat{y} \triangleq E \lbrack Y | w_{1:N}, \alpha, \beta, \delta^2 \rbrack = E\lbrack
\eta^\top \bar{Z} | w_{1:N}, \alpha, \beta, \delta^2 \rbrack.
\label{MedLDA}
\end{eqnarray}

Now, the question underlying the averaging prediction
rule~(\ref{MedLDA}) is how we can devise an appropriate loss
function and constraints to integrate the max-margin concepts of SVR into
latent topic discovery. In the sequel, we present the {\it maximum
entropy discrimination latent Dirichlet allocation} (MedLDA), which is an extension of the PoMEN (i.e., partially observed maximum entropy discrimination Markov networks) \citep{Zhu:08b} framework. PoMEN is
an elegant combination of max-margin learning with structured hidden
variables in Markov networks. The MedLDA is an extension of PoMEN
to learn directed Bayesian networks with latent variables, in particular the latent topic models, which discover latent semantic structures of document collections.

There are two principled choice points in MedLDA according to the prediction rule (\ref{MedLDA}): (1) the distribution
of model parameter $\eta$; and (2) the distribution of latent topic assignment $Z$. Below, we present two MedLDA regression models by using supervised LDA or unsupervised LDA to discover the latent topic assignment $Z$.
Accordingly, we denote these two models as {\it MedLDA}$^{r}_{full}$ and {\it MedLDA}$^{r}_{partial}$.

\subsubsection{Max-Margin Training of sLDA}

For regression, the MedLDA is defined as an integration of a
Bayesian sLDA, where the parameter $\eta$ is sampled from a prior
$p_0(\eta)$, and the $\epsilon$-insensitive support vector
regression (SVR) \citep{Smola:03}. Thus, MedLDA defines a joint
distribution: $p(\theta, \zv, \eta, \yv, \Wv | \alpha, \beta,
\delta^2) = p_0(\eta) p(\theta, \zv, \yv, \Wv | \alpha, \beta, \eta,
\delta^2)$, where the second term is the same as in the sLDA.
Since directly optimizing the log likelihood is intractable, as in sLDA, we optimize its upper
bound. Different from sLDA, $\eta$ is a random variable now. So, we define the variational distribution
$q(\theta, \zv, \eta | \gamma, \phi)$ to approximate the true posterior $p(\theta, \zv, \eta | \alpha, \beta,
\delta^2, \yv, \Wv)$. Then, the upper bound of the negative log-likelihood $- \log p(\yv, \Wv | \alpha, \beta, \delta^2)$ is
\begin{eqnarray}\label{eq:bsLDAbound}
\mathcal{L}^{bs}(q) \triangleq - E_q\lbrack \log p(\theta, \zv, \eta, \yv, \Wv | \alpha, \beta, \delta^2) \rbrack - \mathcal{H}(q(\theta, \zv, \eta )) = KL(q(\eta) \Vert p_0(\eta)) + E_{q(\eta)} \lbrack \mathcal{L}^s \rbrack ,
\end{eqnarray}
\noindent where $KL(p\Vert q) = E_p \lbrack \log(p / q) \rbrack$ is the Kullback-Leibler (KL) divergence.

Thus, the integrated learning problem is defined as: \\[-0.7cm]
\setlength\arraycolsep{2pt}\begin{eqnarray} {\rm
P1(MedLDA}^r_{full}): & & \min_{q, \alpha, \beta, \delta^2, \xi,
\xi^\star}~ \mathcal{L}^{bs}(q) + C \sum_{d=1}^D (\xi_d + \xi_d^\star) \nonumber \\
\mathrm{s.t.}~\forall d: & & \left\{ \begin{array}{rll}
    y_d - E\lbrack \eta^\top \bar{Z}_d \rbrack & \leq \epsilon + \xi_d, & \mu_d \\
    - y_d + E\lbrack \eta^\top \bar{Z}_d \rbrack & \leq \epsilon + \xi_d^\star, & \mu_d^\star \\
    \xi_d & \geq 0, & v_d  \\
    \xi_d^\star & \geq 0, & v_d^\star
    \end{array} \right. \nonumber
\end{eqnarray}
\noindent where $\mu, \mu^\star, v, v^\star$ are lagrange
multipliers; $\xi,\xi^\star$ are slack variables absorbing errors in training
data; and $\epsilon$ is the precision parameter. The constraints in P1 are in the same form as those of P0, but in an expected version because both the latent topic assignments $Z$ and the model parameters $\eta$ are random variables in MedLDA.
Similar as in SVR, the expected constraints correspond to an $\epsilon$-insensitive loss, that is, if the current prediction $\hat{y}$ as in Eq. (\ref{MedLDA}) does not deviate from the target value too much (i.e., less than $\epsilon$), there is no loss; otherwise, a linear loss will be penalized.

The rationale underlying the MedLDA$^r_{full}$ is that: let the current
model be $p(\theta, \zv, \eta, \yv, \Wv | \alpha, \beta, \delta^2)$,
then we want to find a latent topic representation and a model
distribution (as represented by the distribution $q$) which on one
hand tend to predict correctly on the data with a sufficient large
margin, and on the other hand tend to explain the data well (i.e.,
minimizing an variational upper bound of the negative
log-likelihood). The max-margin estimation and topic discovery procedure are coupled together via the constraints, which are defined on the expectations of model parameters $\eta$ and the latent topic representations $Z$. This interplay will yield a topic representation that is more suitable for max-margin learning, as explained below.

{\bf Variational EM-Algorithm}: Solving the constrained problem P1 is generally intractable. Thus, we make
use of mean-field variational methods \citep{Jordan:99} to efficiently obtain an approximate $q$. The basic principle of mean-field variational methods is to form a factorized distribution of the latent variables, parameterized by free variables which are called variational parameters. These parameters are fit so that the KL divergence between the approximate $q$ and the true posterior is small. Variational methods have successfully used in many topic models, as we have presented in Section 2.

As in standard topic models, we assume $q(\theta, \zv, \eta | \gamma, \phi) =
q(\eta) \prod_{d=1}^D q(\theta_d | \gamma_d) \prod_{n=1}^N q(z_{dn}
| \phi_{dn})$, where $\gamma_d$ is a $K$-dimensional vector of
Dirichlet parameters and each $\phi_{dn}$ is a categorical
distribution over $K$ topics. Then, $E\lbrack Z_{dn} \rbrack =
\phi_{dn}$, $E\lbrack \eta^\top \bar{Z}_d \rbrack = E\lbrack \eta
\rbrack^\top (1/N)\sum_{n=1}^N \phi_{dn}$. We can develop an EM
algorithm, which iteratively solves the following two steps: {\it E-step}: infer the posterior distribution of the
hidden variables $\theta$, $Z$, and $\eta$; and {\it M-step}: estimate the unknown model parameters $\alpha$, $\beta$, and $\delta^2$.

The essential difference between MedLDA and sLDA lies in the E-step
to infer the posterior distribution of $\zv$ and $\eta$ because of
the margin constraints in P1. As we shall see in Eq.
(\ref{eq:phiMedLDAr}), these constraints will bias the expected
topic proportions towards the ones that are more suitable for
the supervised prediction tasks. Since the constraints in P1 are not on the
model parameters ($\alpha$, $\beta$, and $\delta^2$), the M-step
is similar to that of the sLDA. We outline the algorithm in Alg.
\ref{alg:coord_descent_regress} and explain it in details below.
Specifically, we formulate a Lagrangian $L$ for P1
{\small \setlength\arraycolsep{1pt} \begin{eqnarray}
L = && \mathcal{L}^{bs}(q) + C \sum_{d=1}^D (\xi_d + \xi_d^\star) - \sum_{d=1}^D \mu_d
(\epsilon + \xi_d - y_d + E\lbrack \eta^\top \bar{Z}_d \rbrack ) -
\sum_{d=1}^D (\mu_d^\star ( \epsilon + \xi_d^\star + y_d - E \lbrack \eta^\top \bar{Z}_d \rbrack ) \nonumber \\
&& + v_d \xi_d + v_d^\star \xi_d^\star ) - \sum_{d=1}^D \sum_{i=1}^N c_{di}(\sum_{j=1}^K \phi_{dij} -1 ), \nonumber
\end{eqnarray}}
\noindent where the last term is due to the normalization condition
$\sum_{j=1}^K \phi_{dij} = 1,~\forall i,d$. Then, the EM procedure alternatively optimize the Lagrangian functional
with respect to each argument.

\begin{enumerate}
\item {\bf E-step}: we infer the posterior distribution of the latent variables $\theta$, $Z$ and $\eta$.
For the variables $\theta$ and $Z$, inferring the posterior distribution is to fit the variational parameters $\gamma$ and $\phi$ because of the mean-field assumption about $q$, but for $\eta$ the optimization is on $q(\eta)$. Specifically, we have the following update rules for different latent variables.

\begin{algorithm}[tb]
   \caption{\footnotesize Variational MedLDA$^r$}
   \label{alg:coord_descent_regress}
\begin{algorithmic}
   \STATE {\bfseries Input:} corpus $\mathcal{D} = \lbrace (\yv, \Wv) \rbrace$, constants
   $C$ and $\epsilon$, and topic number $K$.
   \STATE {\bfseries Output:} Dirichlet parameters $\gamma$, posterior distribution $q(\eta)$, parameters $\alpha$, $\beta$ and $\delta^2$.
   \REPEAT
        \STATE /**** E-Step ****/
        \FOR{$d=1$ {\bfseries to} $D$}
            \STATE Update $\gamma_d$ as in Eq. (\ref{eq:gammaMedLDAr}).
            \FOR{$i=1$ {\bfseries to} $N$}
                \STATE Update $\phi_{di}$ as in Eq. (\ref{eq:phiMedLDAr}).
            \ENDFOR
        \ENDFOR
        \STATE Solve the dual problem D1 to get $q(\eta)$, $\mu$ and $\mu^\star$.
        \STATE /**** M-Step ****/
       \STATE Update $\beta$ using Eq.~(\ref{eq:beta}), and update $\delta^2$ using
       Eq.~(\ref{eq:delta}). $\alpha$ is fixed as $1/K$ times the ones vector.
   \UNTIL{convergence}
\end{algorithmic}
\end{algorithm}

Since the constraints in P1 are not on $\gamma$, optimize $L$ with respect to $\gamma_d$ and we can get the same update formula as in sLDA:
\begin{eqnarray} \label{eq:gammaMedLDAr} \gamma_d \gets
\alpha + \sum_{n=1}^N \phi_{dn}
\end{eqnarray}
Due to the fully factorized assumption of $q$, for each document $d$ and each word $i$, by setting $\partial L / \partial \phi_{di} = 0$, we have:
\setlength\arraycolsep{1pt}\begin{eqnarray}\label{eq:phiMedLDAr}
\phi_{di} \propto \exp \big(  E \lbrack \log \theta | \gamma
\rbrack + E \lbrack \log p(w_{di} | \beta) \rbrack && + \frac{y_d}{N \delta^2} E\lbrack \eta \rbrack - \frac{2 E\lbrack \eta^\top \phi_{d,-i} \eta \rbrack + E\lbrack
\eta \circ \eta \rbrack }{2N^2 \delta^2} \nonumber \\
&& + \frac{E\lbrack \eta \rbrack}{N} ( \mu_d - \mu_d^\star ) \big),
\end{eqnarray}
\noindent where $\phi_{d,-i} = \sum_{n \neq i} \phi_{dn}$; $\eta \circ \eta$ is the element-wise product;
and the result of exponentiating a vector is a vector of the exponentials of
its corresponding components. The first two terms in the exponential
are the same as those in unsupervised LDA.

The essential differences of MedLDA$^r$ from the sLDA lie in the
last three terms in the exponential of $\phi_{di}$. Firstly, the
third and fourth terms are similar to those of sLDA, but in an
expected version since we are learning the distribution $q(\eta)$.
The second-order expectations $E\lbrack \eta^\top \phi_{d,-i} \eta
\rbrack$ and $E\lbrack \eta \circ \eta \rbrack$ mean that the
co-variances of $\eta$ affect the distribution over topics. This
makes our approach significantly different from a point estimation
method, like sLDA, where no expectations or co-variances are
involved in updating $\phi_{di}$. Secondly, the last term is from
the max-margin regression formulation. For a document $d$, which
lies around the decision boundary, i.e., a support vector, either
$\mu_d$ or $\mu_d^\star$ is non-zero, and the last term biases
$\phi_{di}$ towards a distribution that favors a more accurate
prediction on the document. Moreover, the last term is fixed for
words in the document and thus will directly affect the latent
representation of the document, i.e., $\gamma_d$. Therefore, the
latent representation by MedLDA$^r$ is more suitable for max-margin
learning.


Let $A$ be the $D \times K$ matrix whose rows are the vectors $\bar{Z}_d^\top$. Then, we have the following theorem.
\begin{theorem}
For MedLDA, the optimum solution of $q(\eta)$ has the form:
\setlength\arraycolsep{1pt} \begin{eqnarray} q(\eta) =
\frac{ p_0(\eta) }{Z}  \exp \big( \eta^\top \sum_{d=1}^D (\mu_d -
\mu_d^\star + \frac{y_d}{\delta^2}) E\lbrack \bar{Z}_d \rbrack  -
\eta^\top \frac{E\lbrack A^\top A\rbrack}{2\delta^2} \eta \big)
\nonumber
\end{eqnarray}
\noindent where $E\lbrack A^\top A \rbrack = \sum_{d=1}^D E\lbrack
\bar{Z}_d \bar{Z}_d^\top \rbrack$, and $E\lbrack \bar{Z}_d
\bar{Z}_d^\top \rbrack = \frac{1}{N^2} (\sum_{n=1}^N \sum_{m \neq n}
\phi_{dn} \phi_{dm}^\top + \sum_{n=1}^N
{\rm diag}\{\phi_{dn}\} )$. The lagrange multipliers are the solution of the dual problem of P1:
\setlength\arraycolsep{2pt}\begin{eqnarray} {\rm D1}: & &
\max_{\mu, \mu^\star} ~ - \log Z -
\epsilon \sum_{d=1}^D (\mu_d + \mu_d^\star) + \sum_{d=1}^D y_d (\mu_d - \mu_d^\star) \nonumber \\
& & \mathrm{s.t.} ~~\forall d: ~ \mu_d, \mu_d^\star \in \lbrack 0, C
\rbrack. \nonumber
\end{eqnarray}
\end{theorem}
\begin{proof}(sketch)
Set the partial derivative $\partial L / \partial q(\eta)$ equal zero, we can get the solution of $q(\eta)$.
Plugging $q(\eta)$ into $L$, we get the dual problem.
\end{proof}

In MedLDA$^r$, we can choose different priors to introduce some regularization effects. For the standard normal
prior: $p_0(\eta) = \mathcal{N}(0, I)$, we have the corollary:
\begin{corollary}\label{corollary:gaussian}
Assume the prior $p_0(\eta) = \mathcal{N}(0, I)$, then the optimum solution of $q(\eta)$ is
\begin{eqnarray}
q(\eta) = \mathcal{N}\big(\lambda, \Sigma \big),
\end{eqnarray}
\noindent where $\lambda = \Sigma \big(\sum_{d=1}^D (\mu_d - \mu_d^\star +
\frac{y_d}{\delta^2}) E\lbrack \bar{Z}_d \rbrack \big)$ is the mean
and $\Sigma = ( I + 1 / \delta^2 E\lbrack A^\top A \rbrack)^{-1}$ is
a $K \times K$ co-variance matrix. The dual problem of P1 is:
\setlength\arraycolsep{0pt} \begin{eqnarray} & & \max_{\mu,
\mu^\star} ~ - \frac{1}{2} a^\top \Sigma a -
\epsilon \sum_{d=1}^D (\mu_d + \mu_d^\star) + \sum_{d=1}^D y_d (\mu_d - \mu_d^\star) \nonumber \\
& &  \mathrm{s.t.} ~~\forall d: ~ \mu_d, \mu_d^\star \in \lbrack 0,
C \rbrack, \nonumber
\end{eqnarray}
\noindent where $a = \sum_{d=1}^D (\mu_d - \mu_d^\star +
\frac{y_d}{\delta^2}) E\lbrack \bar{Z}_d \rbrack$.
\end{corollary}

In the above Corollary, computation of $\Sigma$ can be achieved robustly through Cholesky decomposition of $\delta^2 I +
E\lbrack A^\top A \rbrack$, an $O(K^3)$ procedure. Another example is the Laplace prior, which can lead to a shrinkage
effect \citep{Zhu:08} that is useful in sparse problems. In this paper, we focus on the normal prior and extension to the Laplace prior can be done similarly as in \citep{Zhu:08}. For the standard normal prior, the dual optimization problem is a QP problem and can be solved with any standard QP solvers, although they may not be so efficient. To leverage recent developments in support vector regression,
we first prove the following corollary:

\begin{corollary}\label{corollary:primal}
Assume the prior $p_0(\eta) = \mathcal{N}(0, I)$, then the mean $\lambda$ of $q(\eta)$ is the optimum solution of the following problem:
\setlength\arraycolsep{1pt}\begin{eqnarray}
\min_{\lambda, \xi, \xi^\star} & & \frac{1}{2} \lambda^\top
\Sigma^{-1} \lambda - \lambda^\top (\sum_{d=1}^D
\frac{y_d}{\delta^2} E\lbrack \bar{Z}_d \rbrack )
+ C \sum_{d=1}^D (\xi_d + \xi_d^\star) \nonumber \\
\mathrm{s.t.}~\forall d: & & \left\{ \begin{array}{rl}
    y_d - \lambda^\top E\lbrack \bar{Z}_d \rbrack & \leq \epsilon + \xi_d \\
    - y_d + \lambda^\top E\lbrack \bar{Z}_d \rbrack & \leq \epsilon + \xi_d^\star \\
    \xi_d, \xi_d^\star & \geq 0
    \end{array} \right. \nonumber
\end{eqnarray}

\end{corollary}

\begin{proof}
See Appendix A for details.
\end{proof}

The above primal form can be re-formulated as a standard SVR problem and solved by using existing
algorithms like SVM-light~\citep{Joachims:99} to get $\lambda$ and the dual parameters $\mu$ and $\mu^\star$. Specifically,
we do Cholesky decomposition $\Sigma^{-1} = U^\top U$, where $U$ is an upper triangular matrix with strict positive diagonal entries. Let $\nu = \sum_{d=1}^D \frac{y_d}{\delta^2} E\lbrack \bar{Z}_d \rbrack$, and we define $\lambda^\prime = U(\lambda - \Sigma \nu)$; $y_d^\prime = y_d - \nu^\top \Sigma E\lbrack \bar{Z}_d \rbrack$; and $\xv_d = (U^{-1})^\top E\lbrack \bar{Z}_d \rbrack$.
Then, the above primal problem in Corollary \ref{corollary:primal} can be re-formulated as the following standard form:
\setlength\arraycolsep{1pt}\begin{eqnarray}
\min_{\lambda^\prime, \xi, \xi^\star} & & \frac{1}{2} \Vert \lambda^\prime \Vert_2^2 + C \sum_{d=1}^D (\xi_d + \xi_d^\star) \nonumber \\
\mathrm{s.t.}~\forall d: & & \left\{ \begin{array}{rl}
    y_d^\prime - (\lambda^\prime)^\top \xv_d & \leq \epsilon + \xi_d \\
    - y_d^\prime + (\lambda^\prime)^\top \xv_d & \leq \epsilon + \xi_d^\star \\
    \xi_d, \xi_d^\star & \geq 0
    \end{array} \right. \nonumber
\end{eqnarray}

\item {\bf M-step}: Now, we estimate the unknown parameters $\alpha,~\beta$, and
$\delta^2$. Here, we assume $\alpha$ is fixed. For $\beta$, the update equations are the same as for sLDA:
\setlength\arraycolsep{0pt}\begin{eqnarray}\label{eq:beta}
~~~~~\beta_{k,w} \propto \sum_{d=1}^D \sum_{n=1}^N 1(w_{dn} = w)
\phi_{dnk},
\end{eqnarray}

For $\delta^2$, this step is similar to that of sLDA but in an expected version. The update rule is:
\setlength\arraycolsep{0pt}\begin{eqnarray}\label{eq:delta}
\delta^2 \gets \frac{1}{D} \big( \yv^\top \yv - 2 \yv^\top E\lbrack
A \rbrack E\lbrack \eta \rbrack + E\lbrack \eta^\top E\lbrack A^\top
A \rbrack \eta \rbrack \big),
\end{eqnarray}
\noindent where $E\lbrack \eta^\top E\lbrack A^\top A \rbrack \eta
\rbrack = {\rm tr}( E\lbrack A^\top A \rbrack E\lbrack \eta
\eta^\top \rbrack)$.

\end{enumerate}

\subsubsection{Max-Margin Learning of LDA for Regression}

In the previous section, we have presented the MedLDA regression model which uses the supervised sLDA to discover the latent topic representations $Z$. The same principle can be applied to perform joint maximum likelihood estimation and max-margin training for the unsupervised LDA \cite{Blei:03}. In this section, we present this MedLDA model, which will be referred to as {\it MedLDA$^r_{partial}$}.

A naive approach to using the unsupervised LDA for supervised prediction tasks, e.g., regression, is a two-step procedure:
(1) using the unsupervised LDA to discover the latent topic representations of documents; and (2) feeding the low-dimensional topic representations into a regression model (e.g., SVR) for training and testing. This de-coupled approach is rather sub-optimal because the side information of documents (e.g., rating scores of movie reviews) is not used in discovering the low-dimensional representations and thus can result in a sub-optimal representation for prediction tasks. Below, we present the MedLDA$^r_{partial}$, which integrates an unsupervised LDA for discovering topics with the SVR for regression. The inter-play between topic discovery and supervised prediction will result in more discriminative latent topic representations, similar as in MedLDA$^r_{full}$.

When the underlying topic model is the unsupervised LDA, the likelihood is $p(\Wv | \alpha, \beta)$ as we have stated. For regression, we apply the $\epsilon$-insensitive support vector regression (SVR) \cite{Smola:03} approach as before. Again, we learn a distribution $q(\eta)$. The prediction rule is the same as in Eq. (\ref{MedLDA}). The integrated learning problem is defined as: \\[-0.7cm]

\setlength\arraycolsep{2pt}\begin{eqnarray} {\rm
P2(MedLDA}^r_{\it partial}): & & \min_{q, q(\eta), \alpha, \beta, \xi,
\xi^\star}~ \mathcal{L}^u(q) + KL(q(\eta) || p_0(\eta)) + C \sum_{d=1}^D (\xi_d + \xi_d^\star) \nonumber \\
\mathrm{s.t.}~\forall d: & & \left\{ \begin{array}{rl}
    y_d - E\lbrack \eta^\top \bar{Z}_d \rbrack & \leq \epsilon + \xi_d \\
    - y_d + E\lbrack \eta^\top \bar{Z}_d \rbrack & \leq \epsilon + \xi_d^\star \\
    \xi_d, \xi_d^\star & \geq 0
    \end{array} \right. ,\nonumber
\end{eqnarray}
\noindent where the $KL$-divergence is a regularizer that bias the estimate of $q(\eta)$ towards the prior. In MedLDA$^r_{\it full}$, this KL-regularizer is implicitly contained in the variational bound $\mathcal{L}^{bs}$ as shown in Eq. (\ref{eq:bsLDAbound}).

{\bf Variational EM-Algorithm}: For MedLDA$^r_{\it partial}$, the constrained optimization problem P2 can be similarly solved
with an EM procedure. Specifically, we make the same independence assumptions about $q$ as in LDA \citep{Blei:03}, that is, we assume that $q(\theta, \zv | \gamma, \phi) = \prod_{d=1}^D q(\theta_d | \gamma_d) \prod_{n=1}^N q(z_{dn}
| \phi_{dn})$, where the variational parameters $\gamma$ and $\phi$ are the same as in MedLDA$^r_{\it full}$. By formulating a Lagrangian $L$ for P2 and iteratively optimizing $L$ over each variable, we can get a variational EM-algorithm that is similar to that of MedLDA$^r_{full}$.
\begin{enumerate}
\item {\bf E-step}: The update rule for $\gamma$ is the same as in MedLDA$^r_{\it full}$. For $\phi$,
by setting $\partial L / \partial \phi_{di} = 0$, we have:
\setlength\arraycolsep{0pt}\begin{eqnarray}\label{eq:phiMedLDAr_simple}
\phi_{di} \propto \exp \big(  E \lbrack \log \theta | \gamma
\rbrack + E \lbrack \log p(w_{di} | \beta) \rbrack + \frac{E\lbrack \eta
\rbrack}{N} ( \mu_d - \mu_d^\star ) \big),
\end{eqnarray}
Compared to the Eq. (\ref{eq:phiMedLDAr}), Eq. (\ref{eq:phiMedLDAr_simple}) is simpler and does not have the complex third and fourth terms of Eq. (\ref{eq:phiMedLDAr}). This simplicity suggests that the latent topic representation is less affected by the max-margin estimation (i.e., the prediction model's parameters).

Set $\partial L / \partial q(\eta) = 0$, then we get:
\setlength\arraycolsep{1pt} \begin{eqnarray} q(\eta) =
\frac{ p_0(\eta) }{Z}  \exp \big( \eta^\top \sum_{d=1}^D (\mu_d -
\mu_d^\star ) E\lbrack \bar{Z}_d \rbrack \big)
\nonumber
\end{eqnarray}
Plugging $q(\eta)$ into $L$, the dual problem D2 is the same as D1.
Again, we can choose different priors to introduce some regularization effects.
For the standard normal prior: $p_0(\eta) = \mathcal{N}(0, I)$, the posterior is also a
normal: $q(\eta) = \mathcal{N}(\lambda, I)$, where
$\lambda = \sum_{d=1}^D (\mu_d - \mu_d^\star) E\lbrack \bar{Z}_d \rbrack $ is the mean.
This identity covariance matrix is much simpler than the covariance matrix $\Sigma$ as in {\it MedLDA}$^r_{full}$, which depends on the latent topic representation $Z$. Since $I$ is independent of $Z$, the prediction model in {\it MedLDA$^r_{partial}$} is less affected by the latent topic representations. Together with the simpler update rule (\ref{eq:phiMedLDAr_simple}), we can conclude that the coupling between the max-margin estimation and the discovery of latent topic representations in {\it MedLDA}$^r_{partial}$ is loser than that of the {\it MedLDA}$^r_{full}$. The loser coupling will lead to inferior empirical performance as we shall see.

For the standard normal prior, the dual problem ${\rm D2}$ is a QP problem:
\setlength\arraycolsep{0pt} \begin{eqnarray} & & \max_{\mu,
\mu^\star} ~ - \frac{1}{2} \Vert \lambda \Vert_2^2  -
\epsilon \sum_{d=1}^D (\mu_d + \mu_d^\star) + \sum_{d=1}^D y_d (\mu_d - \mu_d^\star) \nonumber \\
& &  \mathrm{s.t.} ~~\forall d: ~ \mu_d, \mu_d^\star \in \lbrack 0,
C \rbrack, \nonumber
\end{eqnarray}
\noindent Similarly, we can derive its primal form, which can be reformulated as a standard SVR problem:
\setlength\arraycolsep{1pt}\begin{eqnarray}
\min_{\lambda, \xi, \xi^\star} & & \frac{1}{2} \Vert \lambda \Vert_2^2  - \lambda^\top (\sum_{d=1}^D \frac{y_d}{\delta^2} E\lbrack \bar{Z}_d \rbrack )
+ C \sum_{d=1}^D (\xi_d + \xi_d^\star) \nonumber \\
\mathrm{s.t.}~\forall d: & & \left\{ \begin{array}{rl}
    y_d - \lambda^\top E\lbrack \bar{Z}_d \rbrack & \leq \epsilon + \xi_d \\
    - y_d + \lambda^\top E\lbrack \bar{Z}_d \rbrack & \leq \epsilon + \xi_d^\star \\
    \xi_d, \xi_d^\star & \geq 0 .
    \end{array} \right. \nonumber
\end{eqnarray}
\noindent Now, we can leverage recent developments in support vector
regression to solve either the dual problem or the primal problem.

\item {\bf M-step}: the same as in the MedLDA$^r_{\it full}$.
\end{enumerate}

\section{Maximum Entropy Discrimination LDA for Classification}

In this section, we consider the discrete response variable and present the MedLDA classification model.

\subsection{Learning MedLDA for Classification}

For classification, the response variables $y$ are discrete. For
brevity, we only consider the multi-class classification, where $y
\in \{ 1, \cdots, M \}$. The binary case can be easily defined based
on a binary SVM and the optimization problem can be solved similarly.

For classification, we assume the discriminant function $F$ is
linear, that is, $F(y, z_{1:N}, \eta) = \eta_y^\top \bar{z}$, where
$\bar{z} = 1/N \sum_n z_n$ as in the regression model, $\eta_y$ is a
class-specific $K$-dimensional parameter vector associated with the
class $y$ and $\eta$ is a $MK$-dimensional vector by stacking the
elements of $\eta_y$. Equivalently, $F$ can be written as $F(y,
z_{1:N}, \eta) = \eta^\top \fv(y, \bar{z})$, where $\fv(y, \bar{z})$
is a feature vector whose components from $(y-1)K+1$ to $yK$ are
those of the vector $\bar{z}$ and all the others are 0. From each
single $F$, a prediction rule can be derived as in SVM. Here, we
consider the general case to learn a distribution of $q(\eta)$ and
for prediction, we take the average over all the possible models and
the latent topics:

\begin{equation}\label{eq:classifier}
y^\star = {\rm arg}\max_{y} E \lbrack \eta^\top \fv(y, \bar{Z}) |
\alpha, \beta \rbrack.
\end{equation}

Now, the problem is to learn an optimal set of parameters $\alpha,
\beta$ and distribution $q(\eta)$. Below, we present the MedLDA classification model.
In principle, we can develop two variants of MedLDA classification models, which use
the supervised sLDA \citep{Wang:09} and the unsupervised LDA to discover latent topics
as in the regression case. However, for the case of using supervised sLDA for classification, it is impossible to derive a
dual formulation of its optimization problem because of the normalized non-Gaussian prediction model \citep{Blei:07,Wang:09}. Here, we consider the case where we use the unsupervised LDA as the underlying topic model to discover the latent topic representation $Z$. As we shall see, the MedLDA classification model can be easily
learned by using existing SVM solvers to optimize its dual optimization problem.

\subsubsection{Max-Margin Learning of LDA for Classification}

As we have stated, the supervised sLDA model has a normalization factor that makes the learning
generally intractable, except for some special cases like the normal
distribution as in the regression case. In \citep{Blei:07,Wang:09}, variational methods or
high-order Taylor expansion is applied to approximate the normalization factor in classification model.
In our max-margin formulation, since our target is to directly minimize a hinge loss,
we do not need a normalized distribution model for the response variables $Y$.
Instead, we define a {\it partially} generative model on $(\theta, \zv,\Wv)$ only as in the
unsupervised LDA, and for the classification (i.e., from $Z$ to $Y$), we apply the
max-margin principle, which does not require a normalized distribution.
Thus, in this case, the likelihood of the corpus $\mathcal{D}$ is  $p(\Wv | \alpha, \beta)$.

Similar as in the MedLDA$^r_{\it partial}$ regression model, we define the integrated latent
topic discovery and multi-class classification model as follows:
\setlength\arraycolsep{-1.25pt} \begin{eqnarray} {\rm
P3(MedLDA^c)}: && \min_{q, q(\eta), \alpha, \beta,
\xi} \mathcal{L}^u(q) + KL(q(\eta) || p_0(\eta)) + C \sum_{d=1}^D \xi_d \nonumber \\
\mathrm{s.t.} ~\forall d,~y \neq y_d: &&~~~~E\lbrack \eta^\top
\Delta \fv_d(y) \rbrack \geq 1 - \xi_d; ~\xi_d \geq 0, \nonumber
\end{eqnarray}
\noindent where $q(\theta, \zv | \gamma, \phi)$ is a variational
distribution; $\mathcal{L}^u(q)$ is a variational upper bound of $- \log p(\Wv | \alpha, \beta)$; $\Delta
\fv_d(y) = \fv(y_d, \bar{Z}_d) - \fv(y, \bar{Z}_d)$, and $\xi$ are
slack variables. $E\lbrack \eta^\top \Delta \fv_d(y) \rbrack$ is the
``{\it expected} margin'' by which the true label $y_d$ is favored
over a prediction $y$. These margin constraints make MedLDA$^c$ fundamentally different
from the mixture of conditional max-entropy models \citep{Pavlov:03},
where constraints are based on moment matching, i.e., empirical
expectations of features are equal to their model expectations.

The rationale underlying the MedLDA$^c$ is similar to that of the
MedLDA$^r$, that is, we want to find a latent topic representation
$q(\theta, \zv | \gamma, \phi)$ and a parameter distribution
$q(\eta)$ which on one hand tend to predict as accurate as possible
on training data, while on the other hand tend to explain the data
well. The KL-divergence term in P3 is a regularizer of the distribution $q(\eta)$.

\subsection{Variational EM-Algorithm}

As in MedLDA$^r$, we can develop a similar variational EM algorithm.
Specifically, we assume that $q$ is fully factorized, as in the
standard unsupervised LDA. Then, $E\lbrack \eta^\top \fv(y,
\bar{Z}_d) \rbrack = E\lbrack \eta \rbrack^\top \fv(y,
1/N\sum_{n=1}^N \phi_{dn})$. We formulate the Lagrangian $L$ of P3:
{\small \setlength\arraycolsep{1pt}\begin{eqnarray}
L  = \mathcal{L}(q) + KL(q(\eta) || p_0(\eta)) + C \sum_{d=1}^D
\xi_d && - \sum_{d=1}^D v_d \xi_d  - \sum_{d=1}^D \sum_{y \neq y_d}
\mu_d(y) ( E\lbrack \eta^\top \Delta \fv_d(y) \rbrack + \xi_d - 1) \nonumber \\
&& - \sum_{d=1}^D \sum_{i=1}^N c_{di}(\sum_{j=1}^K \phi_{dij} -1 ), \nonumber
\end{eqnarray}}
\noindent where the last term is from the normalization condition
$\sum_{j=1}^K \phi_{dij} = 1,~\forall i,d$. The EM-algorithm iteratively optimizes $L$ w.r.t $\gamma$, $\phi$, $q(\eta)$ and
$\beta$. Since the constraints in P3 are not on $\gamma$ or $\beta$,
their update rules are the same as in MedLDA$^r_{\it full}$ and we omit the
details here. We explain the optimization of $L$ over $\phi$ and
$q(\eta)$ and show the insights of the max-margin topic model:
\begin{enumerate}
\item {\bf Optimize $L$ over $\phi$}: again, since $q$ is fully factorized, we can perform the optimization on each document
separately. Set $\partial L / \partial \phi_{di} = 0$, then we have:
\setlength\arraycolsep{1pt} \begin{eqnarray} \phi_{di}
\propto \exp \big( && E \lbrack \log \theta | \gamma \rbrack + E
\lbrack \log p(w_{di} | \beta) \rbrack + \frac{1}{N} \sum_{y \neq y_d} \mu_d(y) E\lbrack \eta_{y_d} - \eta_y
\rbrack \big). \label{eq:phiMedLDA}
\end{eqnarray}

The first two terms in Eq. (\ref{eq:phiMedLDA}) are the same as in
the unsupervised LDA and the last term is due to the max-margin
formulation of P3 and reflects our intuition that the discovered
latent topic representation is influenced by the max-margin
estimation. For those examples that are around the decision
boundary, i.e., support vectors, some of the lagrange multipliers
are non-zero and thus the last term acts as a regularizer that
biases the model towards discovering a latent representation that
tends to make more accurate prediction on these difficult examples.
Moreover, this term is fixed for words in the document and thus will
directly affect the latent representation of the document (i.e.,
$\gamma_d$) and will yield a discriminative latent representation,
as we shall see in Section~\ref{sec:result}, which is more suitable
for the classification task.

\item {\bf Optimize $L$ over $q(\eta)$}: Similar as in the regression model, we have the following optimum solution.

\begin{corollary}
The optimum solution $q(\eta)$ of MedLDA$^c$ has the form:
\begin{eqnarray}
q(\eta) = \frac{1}{Z} p_0(\eta) \exp \Big( \eta^\top (\sum_{d=1}^D \sum_{y \neq y_d} \mu_d(y) E\lbrack \Delta \fv_d(y) \rbrack) \Big),
\end{eqnarray}
\noindent The lagrange multipliers $\mu$ are the optimum solution of the dual problem:
\begin{eqnarray}
{\rm D3}: & & \max_{\mu} ~ - \log Z + \sum_{d=1}^D \sum_{y \neq y_d} \mu_d(y) \nonumber \\
& & \mathrm{s.t.} ~~\forall d: ~ \sum_{y \neq y_d} \mu_d(y) \in
\lbrack 0, C \rbrack, \nonumber
\end{eqnarray}
\end{corollary}


Again, we can choose different priors in MedLDA$^c$ for different
regularization effects. We consider the normal prior in this paper.
For the standard normal prior $p_0(\eta) = \mathcal{N}(0, I)$, we
can get: $q(\eta)$ is a normal with a shifted mean, i.e., $q(\eta) =
\mathcal{N}(\lambda, I)$, where $\lambda = \sum_{d=1}^D \sum_{y \neq y_d} \mu_d(y) E\lbrack \Delta \fv_d(y) \rbrack$, and the dual problem ${\rm D3}$ is the same as the dual problem of a standard multi-class SVM that can be solved using existing SVM methods \citep{Crammer:01}:
\setlength\arraycolsep{1pt} \begin{eqnarray} & & \max_{\mu}
~ - \frac{1}{2} \Vert \sum_{d=1}^D \sum_{y \neq y_d}
\mu_d(y) E\lbrack \Delta \fv_d(y) \rbrack \Vert_2^2 + \sum_{d=1}^D \sum_{y \neq y_d} \mu_d(y) \nonumber \\
& & \mathrm{s.t.} ~~\forall d: ~ \sum_{y \neq y_d} \mu_d(y) \in
\lbrack 0, C \rbrack. \nonumber
\end{eqnarray}

\end{enumerate}

\section{MedTM: a general framework}

We have presented MedLDA, which integrates the max-margin principle
with an {\it underlying} LDA model, which can be supervised or
unsupervised, for discovering predictive latent topic representations of documents.
The same principle can be applied to other generative topic models, such as the correlated topic models (CTMs) \citep{Blei:nips05}, as well as undirected random fields, such
as the exponential family harmoniums (EFH) \citep{Welling:nips04}.

Formally, the max-entropy discrimination topic models (MedTM) can be generally defined as:
\setlength\arraycolsep{0pt} \begin{eqnarray}
 {\rm P(MedTM}): \min_{q(H), q(\Upsilon),\Psi,\xi} && \mathcal{L}(q(H)) + KL(q(\Upsilon) \Vert p_0(\Upsilon)) + U(\xi) \nonumber \\
   \mathrm{s.t.}~ \textrm{\emph{expected}} &&~ \textrm{margin constraints,} \nonumber 
\end{eqnarray}
\noindent where $H$ are hidden variables (e.g., $(\theta,\zv)$ in
LDA); $\Upsilon$ are the parameters of the model pertaining to the
prediction task (e.g., $\eta$ in sLDA); $\Psi$ are the parameters of
the underlying topic model (e.g., the Dirichlet parameter $\alpha$);
and $\mathcal{L}$ is a variational upper bound of the negative log
likelihood associated with the underlying topic model. $U$ is a
convex function over slack variables. For the general MedTM model,
we can develop a similar variational EM-algorithm as for the MedLDA.
Note that $\Upsilon$ can be a part of $H$. For example, the
underlying topic model of MedLDA$^r$ is a Bayesian sLDA. In this
case, $H=(\theta, \zv, \eta)$, $\Upsilon = \emptyset$ and the term
$KL(q(\eta) \Vert p_0(\eta))$ is contained in its $\mathcal{L}$.

Finally, based on the recent extension of {\it maximum entropy discrimination} (MED) \citep{Jaakkola:99} to the structured prediction setting \citep{Zhu:08b}, the basic principle of MedLDA can be similarly extended to perform structured prediction, where multiple response variables are predicted simultaneously and thus their mutual dependencies can be exploited to achieve global consistent and optimal predictions. Likelihood based structured prediction latent topic models have been developed in different scenarios, such as image annotation \citep{He:08} and statistical machine translation \citep{Zhao:06}. The extension of MedLDA to structured prediction setting could provide a promising alternative for such problems.

\section{Experiments}\label{sec:result}

In this section, we provide qualitative as well as quantitative
evaluation of MedLDA on text modeling, classification and
regression.

\subsection{Text Modeling}

\begin{figure}
\vspace{-1cm}\centerline{\subfigure{\includegraphics[width=402pt]{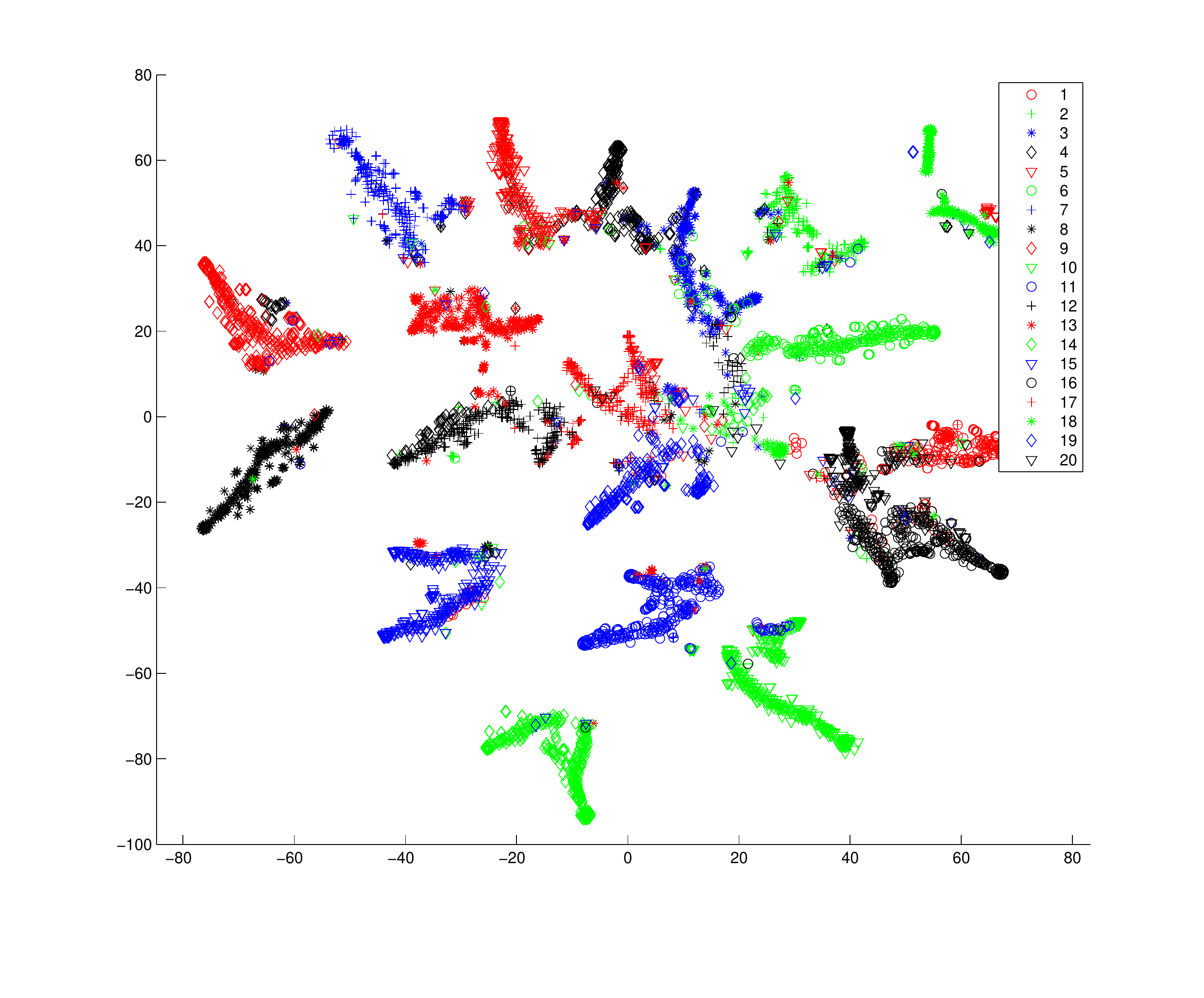}}}
\vspace{-1.8cm}\centerline{\subfigure{\includegraphics[width=402pt]{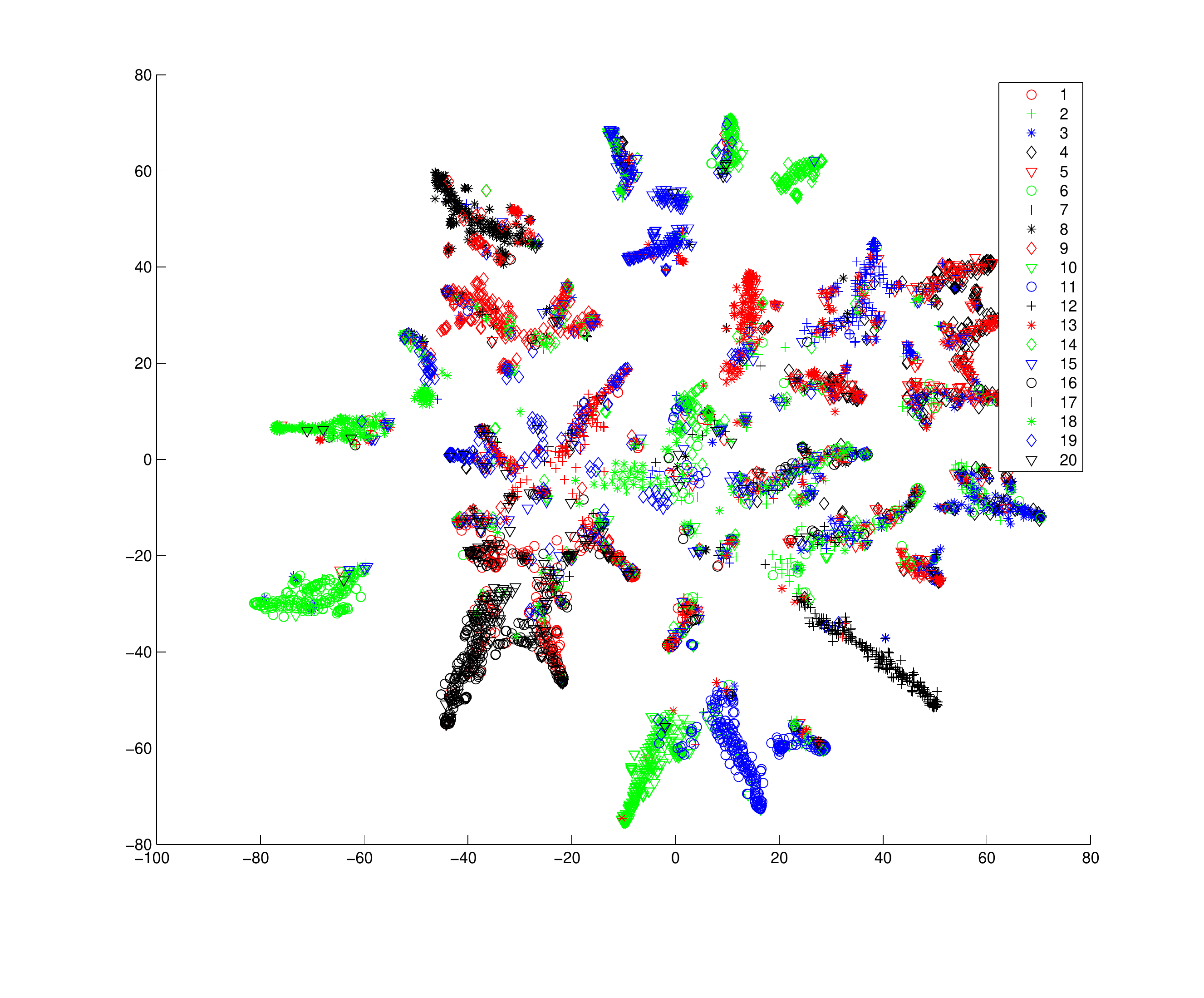}}}%
\vspace{-1.8cm}\caption{t-SNE 2D embedding of the
topic representation by: MedLDA$^c$ (above) and the unsupervised LDA
(below).} \label{fig_embedding}
\end{figure}

\begin{figure}[t]
 \scalebox{.55}{
            \centering
            \begin{tabular}{|p{1in}|c|c|c||c|c|c||cc|}
            \hline
            Class &  \multicolumn{3}{c||}{MedLDA} & \multicolumn{3}{c||}{LDA}&\multicolumn{2}{c|}{Average $\theta$ per class}\\
            \hline
        \multirow{12}{*}{comp.graphics}&&&&&&&&\\
        &&&&&&&\multicolumn{2}{c|}{\multirow{12}{*}{\includegraphics[width=.76\columnwidth,height=.36\columnwidth]{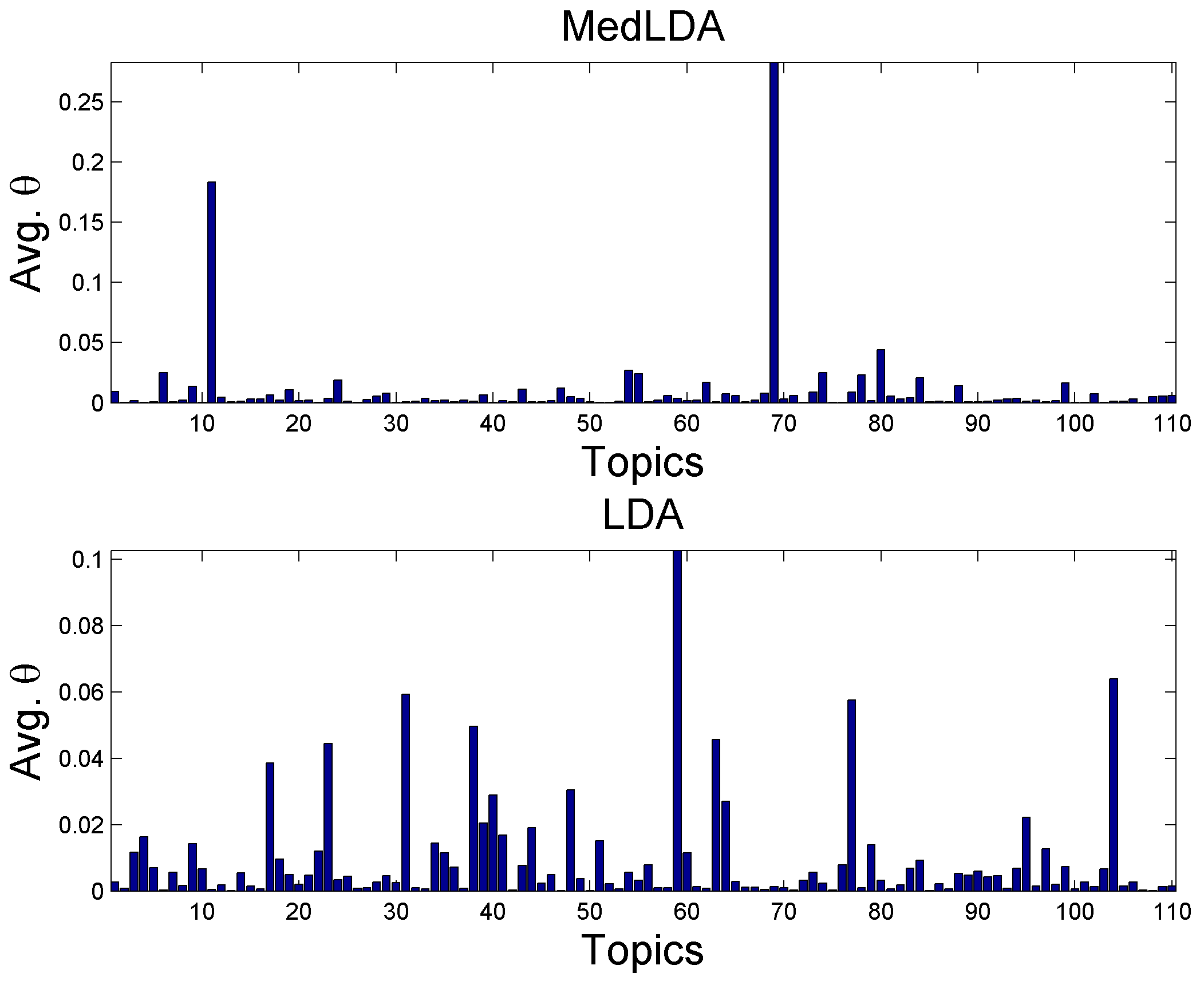}}}\\
        & T $69$ & T $11$ & T $80$ & T $59$ & T $104$ & T $31$ &&\\
        \cline{2-7}
        & image & graphics & db & image & ftp & card &&\\
        & jpeg & image & key & jpeg & pub & monitor &&\\
        & gif & data & chip & color & graphics & dos &&\\
        & file & ftp & encryption & file & mail & video &&\\
        & color & software & clipper & gif & version & apple &&\\
        & files & pub & system & images & tar & windows &&\\
        & bit & mail & government & format & file & drivers &&\\
        & images & package & keys & bit & information & vga &&\\
        & format & fax & law & files & send & cards &&\\
        & program & images & escrow & display & server & graphics &&\\
        \hline
        \multirow{12}{*}{sci.electronics}&&&&&&&&\\
        &&&&&&&\multicolumn{2}{c|}{\multirow{12}{*}{\includegraphics[width=.75\columnwidth,height=.36\columnwidth]{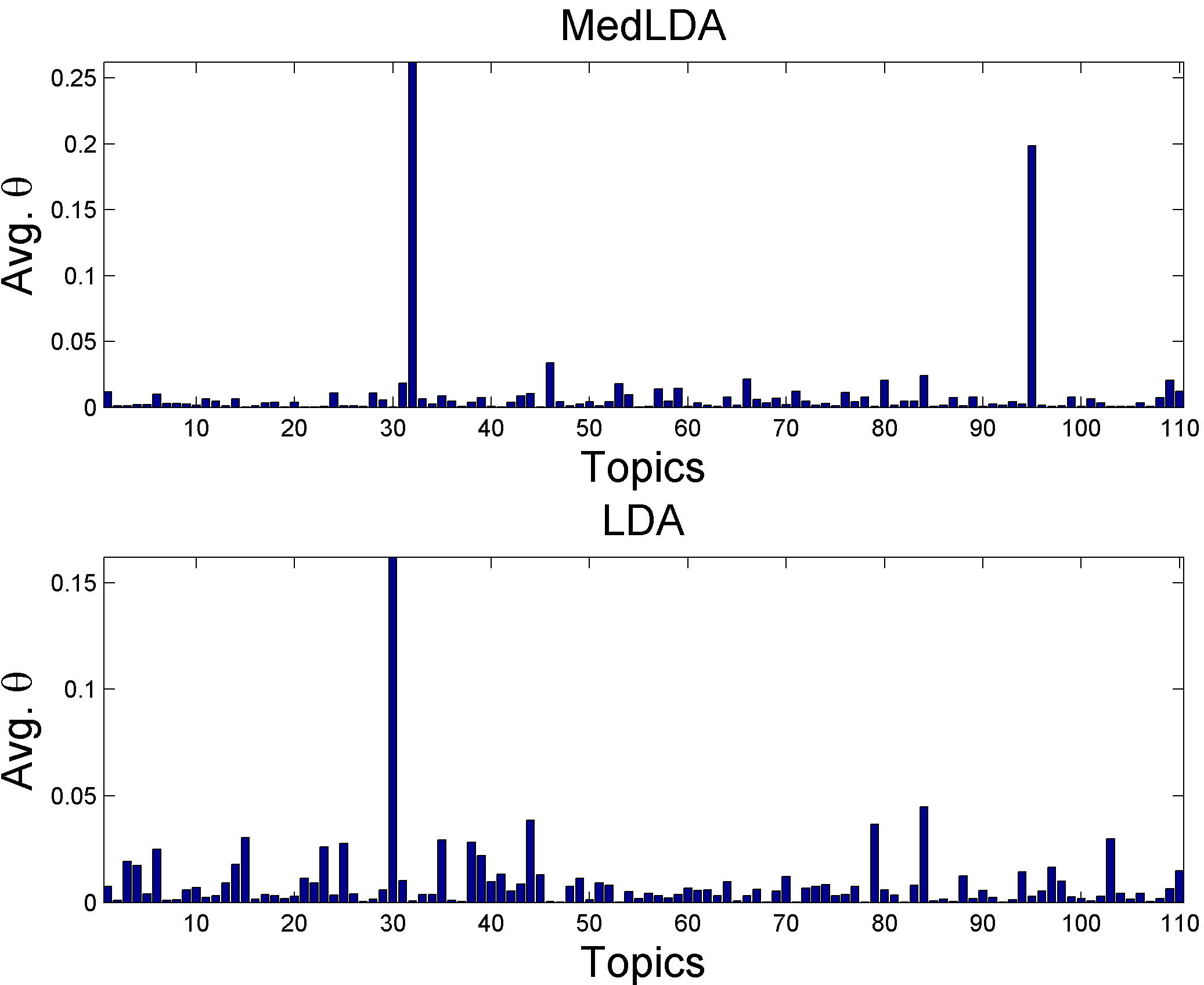}}}\\
        & T $32$ & T $95$ & T $46$ & T $30$ & T $84$ & T $44$ &&\\
        \cline{2-7}
        & ground & audio & source & power & water & sale &&\\
        & wire & output & rs & ground & energy & price &&\\
        & power & input & time & wire & air & offer &&\\
        & wiring & signal & john & circuit & nuclear & shipping &&\\
        & don & chip & cycle & supply & loop & sell &&\\
        & current & high & low & voltage & hot & interested &&\\
        & circuit & data & dixie & current & cold & mail &&\\
        & neutral & mhz & dog & wiring & cooling & condition &&\\
        & writes & time & weeks & signal & heat & email &&\\
        & work & good & face & cable & temperature & cd &&\\
        \hline
        \multirow{12}{*}{politics.mideast}&&&&&&&&\\
        &&&&&&&\multicolumn{2}{c|}{\multirow{12}{*}{\includegraphics[width=.75\columnwidth,height=.36\columnwidth]{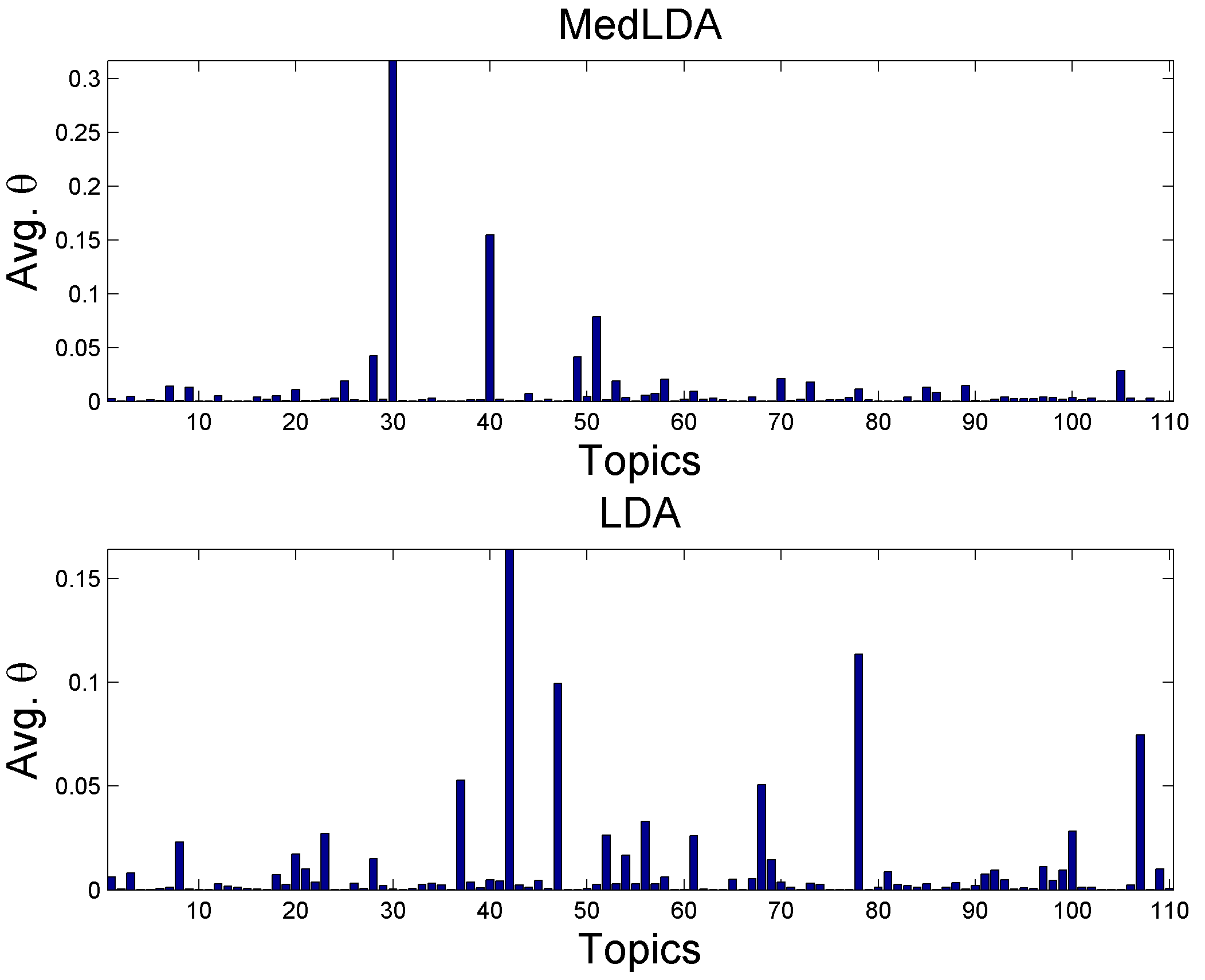}}}\\
        & T $30$ & T $40$ & T $51$ & T $42$ & T $78$ & T $47$ &&\\
        \cline{2-7}
        & israel & turkish & israel & israel & jews & armenian &&\\
        & israeli & armenian & lebanese & israeli & jewish & turkish &&\\
        & jews & armenians & israeli & peace & israel & armenians &&\\
        & arab & armenia & lebanon & writes & israeli & armenia &&\\
        & writes & people & people & article & arab & turks &&\\
        & people & turks & attacks & arab & people & genocide &&\\
        & article & greek & soldiers  & war & arabs & russian &&\\
        & jewish & turkey & villages & lebanese & center & soviet &&\\
        & state & government & peace & lebanon & jew & people &&\\
        & rights & soviet & writes & people & nazi & muslim &&\\
        \hline
        \multirow{12}{*}{misc.forsale}&&&&&&&&\\
        &&&&&&&\multicolumn{2}{c|}{\multirow{12}{*}{\includegraphics[width=.75\columnwidth,height=.36\columnwidth]{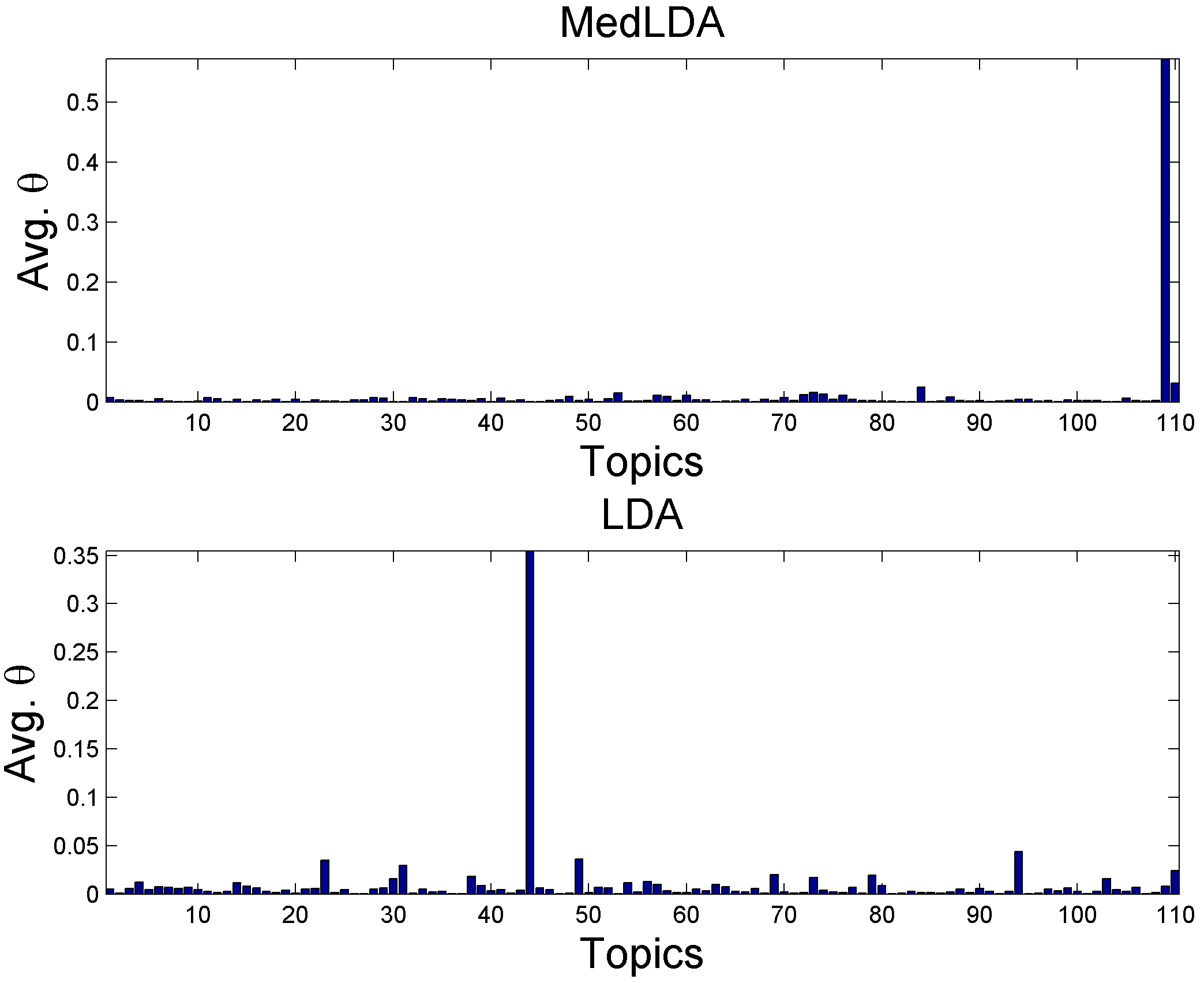}}}\\
        & T $109$ & T $110$ & T $84$ & T $44$ & T $94$ & T $49$ &&\\
        \cline{2-7}
        & sale & drive & mac & sale & don & drive &&\\
        & price & scsi & apple & price & mail & scsi &&\\
        & shipping & mb & monitor & offer & call & disk &&\\
        & offer & drives & bit & shipping & package & hard &&\\
        & mail & controller & mhz & sell & writes & mb &&\\
        & condition & disk & card & interested & send & drives &&\\
        & interested & ide & video & mail & number & ide &&\\
        & sell & hard & speed & condition & ve & controller &&\\
        & email & bus & memory & email & hotel & floppy &&\\
        & dos & system & system & cd & credit & system &&\\
        \hline
        \end{tabular}
            }
    \caption{Top topics under each class as discovered by the MedLDA and LDA models}\vspace{-0.3cm}\label{fig:topics}
\end{figure}

We study text modeling of the MedLDA on the 20 Newsgroups data set
with a standard list of stop
words\footnote{http://mallet.cs.umass.edu/} removed. The data set
contains postings in 20 related categories. We compare with the
standard unsupervised LDA. We fit the dataset to a 110-topic
MedLDA$^c$ model, which explores the supervised category
information, and a 110-topic unsupervised LDA.

Figure~\ref{fig_embedding} shows the 2D embedding of the expected
topic proportions of MedLDA$^c$ and LDA by using the t-SNE
stochastic neighborhood embedding \citep{Hinton:08}, where each dot
represents a document and color-shape pairs represent class labels.
Obviously, the max-margin based MedLDA$^c$ produces a better
grouping and separation of the documents in different categories. In
contrast, the unsupervised LDA does not produce a well separated
embedding, and documents in different categories tend to mix
together. A similar embedding was presented in \citep{Simon:08},
where the transformation matrix in their model is pre-designed. The
results of MedLDA$^c$ in Figure~\ref{fig_embedding} are {\it
automatically} learned.

It is also interesting to examine the discovered topics and their
association with class labels. In Figure~\ref{fig:topics} we show
the top topics in four classes as discovered by both MedLDA and LDA.
Moreover, we depict the per-class distribution over topics for each
model. This distribution is computed by averaging the expected
latent representation of the documents in each class. We can see
that MedLDA yields sharper, sparser and fast decaying per-class
distributions over topics which have a better discrimination power.
This behavior is in fact due to the regularization effect enforced
over $\phi$ as shown in Eq.~(\ref{eq:phiMedLDA}). On the other hand,
LDA seems to discover topics that model the fine details of
documents with no regard to their discrimination power (i.e. it
discovers different variations of the same topic which results in a
flat per-class distribution over topics). For instance, in the class
comp.graphics, MedLDA mainly models documents in this class using
two salient, discriminative topics (T69 and T11) whereas LDA results
in a much flatter distribution. Moreover, in the cases where LDA and
MedLDA discover comparably the same set of topics in a given class
(like politics.mideast and misc.forsale), MedLDA results in a
sharper low dimensional representation.

\subsection{Prediction Accuracy}

In this subsection, we provide a quantitative evaluation of the
MedLDA on prediction performance.

\subsubsection{Classification}

We perform binary and multi-class classification on the 20 Newsgroup
data set. To obtain a baseline, we first fit all the data to an LDA
model, and then use the latent representation of the
training\footnote{We use the training/testing split in:\\
http://people.csail.mit.edu/jrennie/20Newsgroups/} documents as
features to build a binary/multi-class SVM classifier. We denote
this baseline by LDA+SVM. For a model $\mathcal{M}$, we evaluate its
performance using the relative improvement ratio, i.e.,
$\frac{precision(\mathcal{M})~ -~
precision(LDA+SVM)}{precision(LDA+SVM)}$.

Note that since DiscLDA \citep{Simon:08} is using the Gibbs sampling for inference, which is
slightly different from the variational methods as in MedLDA and sLDA \citep{Blei:07,Wang:09},
we build the baseline model of LDA+SVM with both variational inference and Gibbs sampling. The relative improvement
ratio of each model is computed against the baseline with the same inference method.

{\bf Binary Classification}: As in \citep{Simon:08}, the binary
classification is to distinguish postings of the newsgroup
\emph{alt.atheism} and the postings of the group
\emph{talk.religion.misc}. We compare MedLDA$^c$ with sLDA, DiscLDA
and LDA+SVM. For sLDA, the extension to perform multi-class classification was presented by \cite{Wang:09}, we will compare with it in the multi-class classification setting. Here, for binary case, we fit an sLDA regression model using the binary representation (0/1) of the classes, and use a threshold 0.5 to make prediction. For MedLDA$^c$,
to see whether a second-stage max-margin classifier can improve the
performance, we also build a method {\it MedLDA+SVM}, similar to
LDA+SVM. For all the above methods that utilize the class label
information, they are fit {\it ONLY} on the training data. 

\begin{figure}
\centerline{
\subfigure[]{\includegraphics[width=220pt]{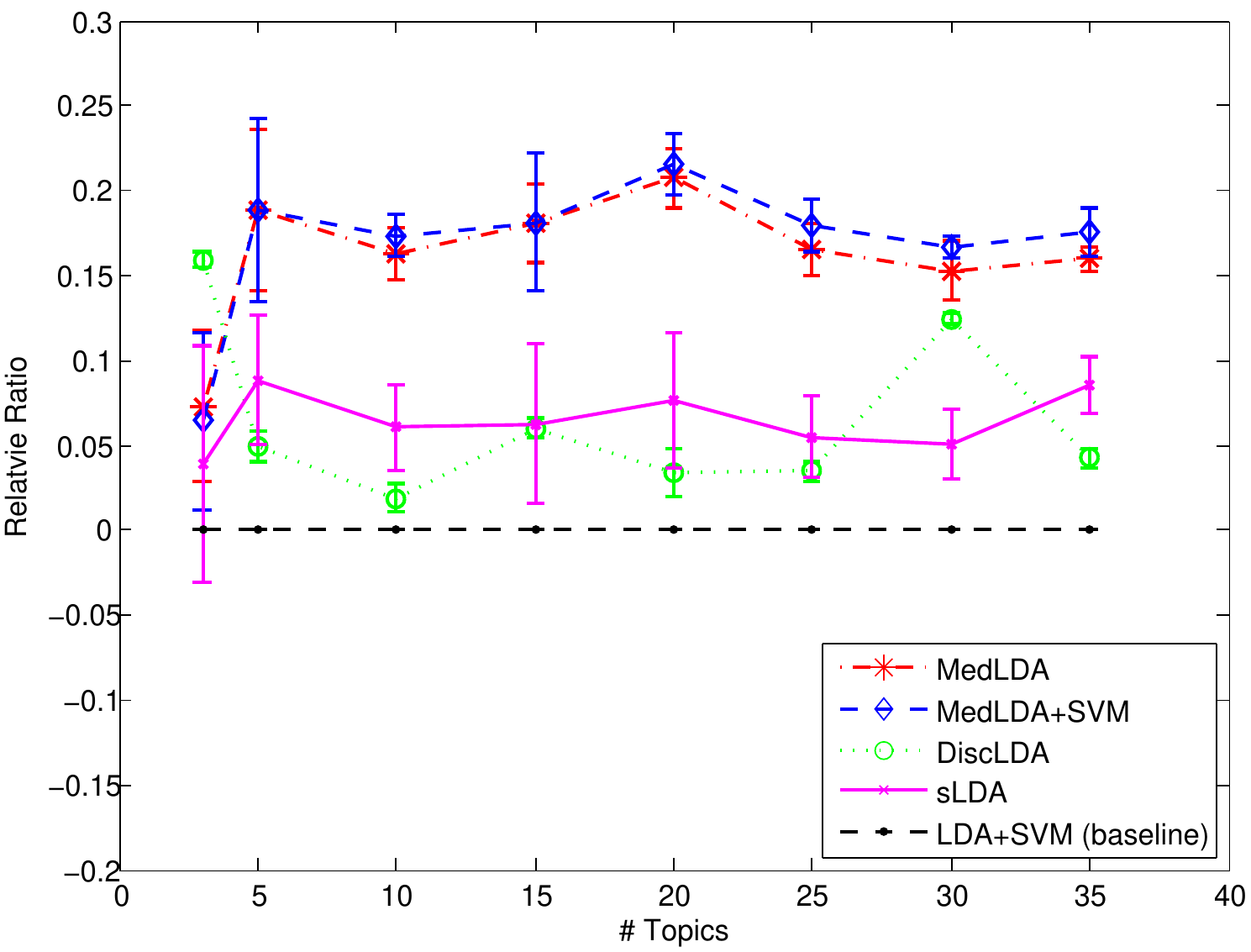}\label{fig_20newsgroup}}
\subfigure[]{\includegraphics[width=220pt]{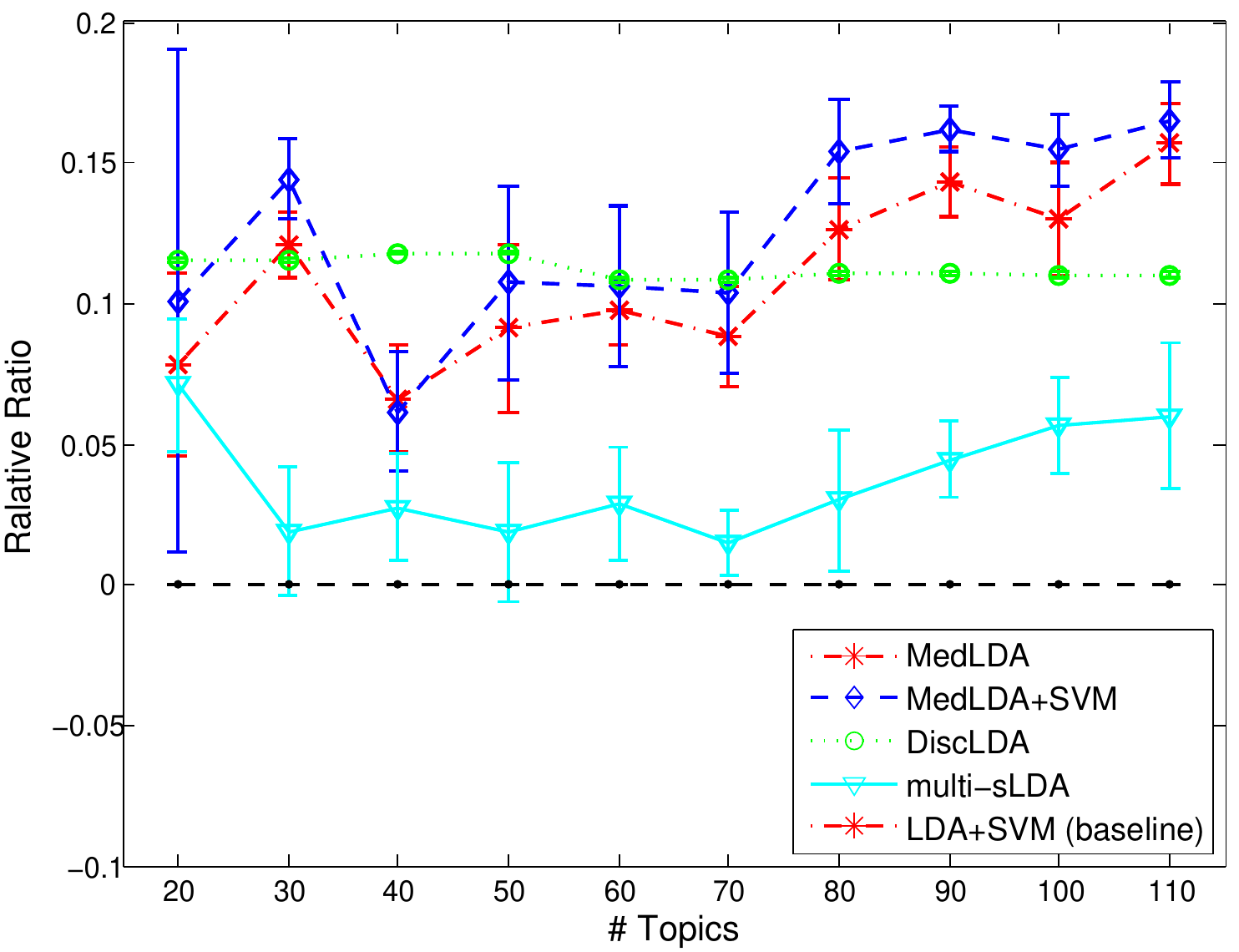}\label{fig_20newsgroup_multiclass}}}
\caption{\footnotesize Relative improvement ratio
against LDA+SVM for: (a) binary and (b) multi-class
classification.}\vspace{-0.6cm}
\end{figure}

We use the SVM-light \citep{Joachims:99} to build SVM classifiers and
to estimate $q(\eta)$ in MedLDA$^c$. The parameter $C$ is chosen via
5 fold cross-validation during the training from $\lbrace k^2: k =
1, \cdots ,8 \rbrace$. For each model, we run the experiments for 5
times and take the average as the final results. 
The relative improvement ratios of different models with respect to topic
numbers are shown in Figure~\ref{fig_20newsgroup}. For the DiscLDA \citep{Simon:08},
the number of topics is set by the equation $2K_0 + K_1$, where $K_0$ is the number of topics per class and $K_1$ is the number of topics shared by all categories. As in \citep{Simon:08}, $K_1 = 2 K_0$. Here, we set $K_0 = 1, \cdots, 8, 10$ and align the
results with those of MedLDA and sLDA that have the closest topic numbers.

We can see that the max-margin based MedLDA$^c$ works better than
sLDA, DiscLDA and the two-step method of LDA+SVM.
Since MedLDA$^c$ integrates the max-margin principle in its
training, the combination of MedLDA and SVM does not yield
additional benefits on this task. We believe that the slight
differences between MedLDA and MedLDA+SVM 
are due to tuning of the regularization parameters. For efficiency,
we do not change the regularization constant $C$ during training
MedLDA$^c$. The performance would be improved if we select a good
$C$ in different iterations because the data representation is
changing.

{\bf Multi-class Classification}: We perform multi-class
classification on 20 Newsgroups with all the categories. We compare
MedLDA$^c$ with MedLDA+SVM, LDA+SVM, multi-class sLDA (multi-sLDA) \citep{Wang:09}, and DiscLDA. We use the
SVM$^{struct}$ package\footnote{http://svmlight.joachims.org/svm\_multiclass.html}
with a 0/1 loss to solve the sub-step of learning $q(\eta)$ and
build the SVM classifiers for LDA+SVM and MedLDA+SVM. The results
are shown in Figure~\ref{fig_20newsgroup_multiclass}. For DiscLDA, we use the same equation as in \citep{Simon:08} to set the number of topics and set $K_0 = 1, \cdots, 5$. Again, we need to align the results with those of MedLDA based on
the closest topic number criterion. We can see that all the supervised topic models discover more predictive topics
for classification, and the max-margin based MedLDA$^c$ can achieve
significant improvements with an appropriate number (e.g., $\geq
80$) of topics. Again, we believe that the slight difference between
MedLDA$^c$ and MedLDA+SVM is due to parameter tuning.

\subsubsection{Regression}

We evaluate the MedLDA$^r$ model on the movie review data set. As in
\citep{Blei:07}, we take logs of the response values to make them
approximately normal. We compare MedLDA$^r$ with the unsupervised
LDA and sLDA. As we have stated, the underlying topic model in
MedLDA$^r$ can be a LDA or a sLDA. We have implemented both, as
denoted by {\it MedLDA (partial)} and {\it MedLDA (full)},
respectively. For LDA, we use its low dimensional representation of
documents as input features to a linear SVR and denote this method
by {\it LDA+SVR}. The evaluation criterion is predictive R$^2$
(pR$^2$) as defined in \citep{Blei:07}.

\begin{figure}
\centerline{
\subfigure{\includegraphics[width=215pt]{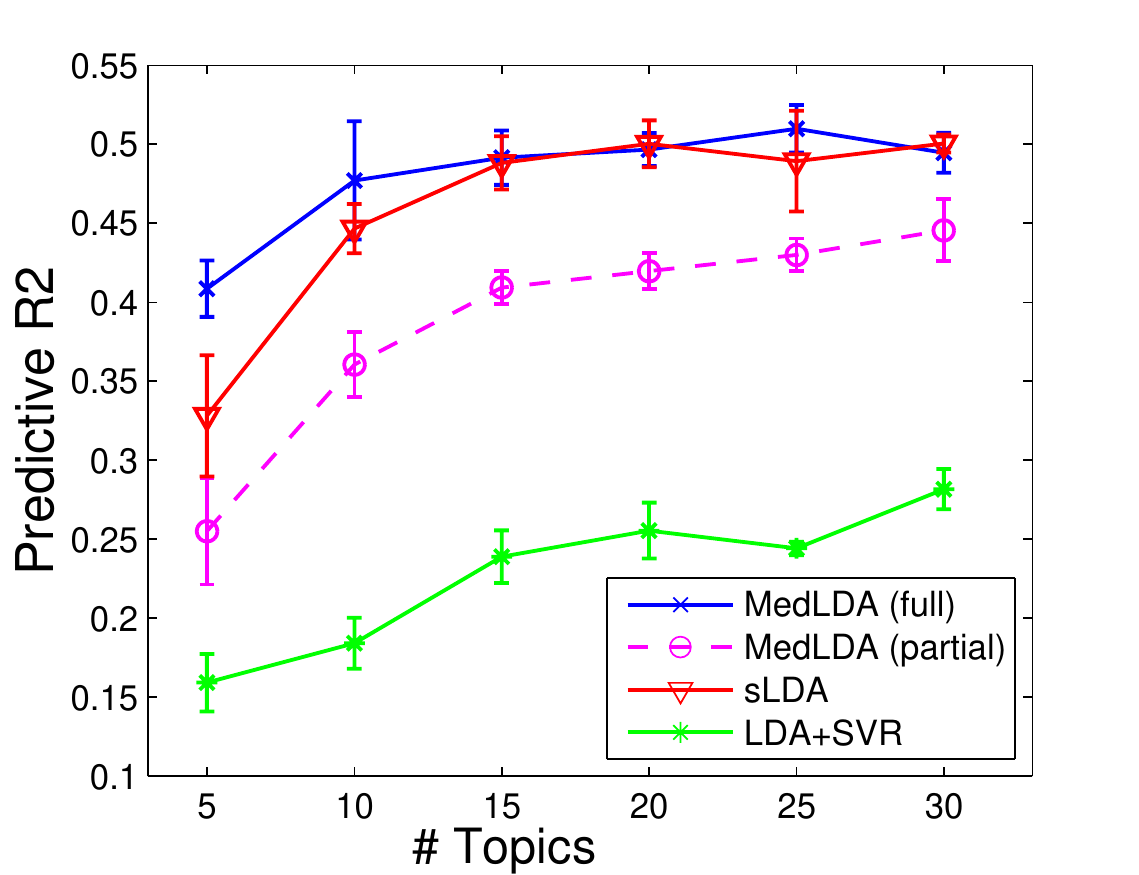}}
\subfigure{\includegraphics[width=215pt]{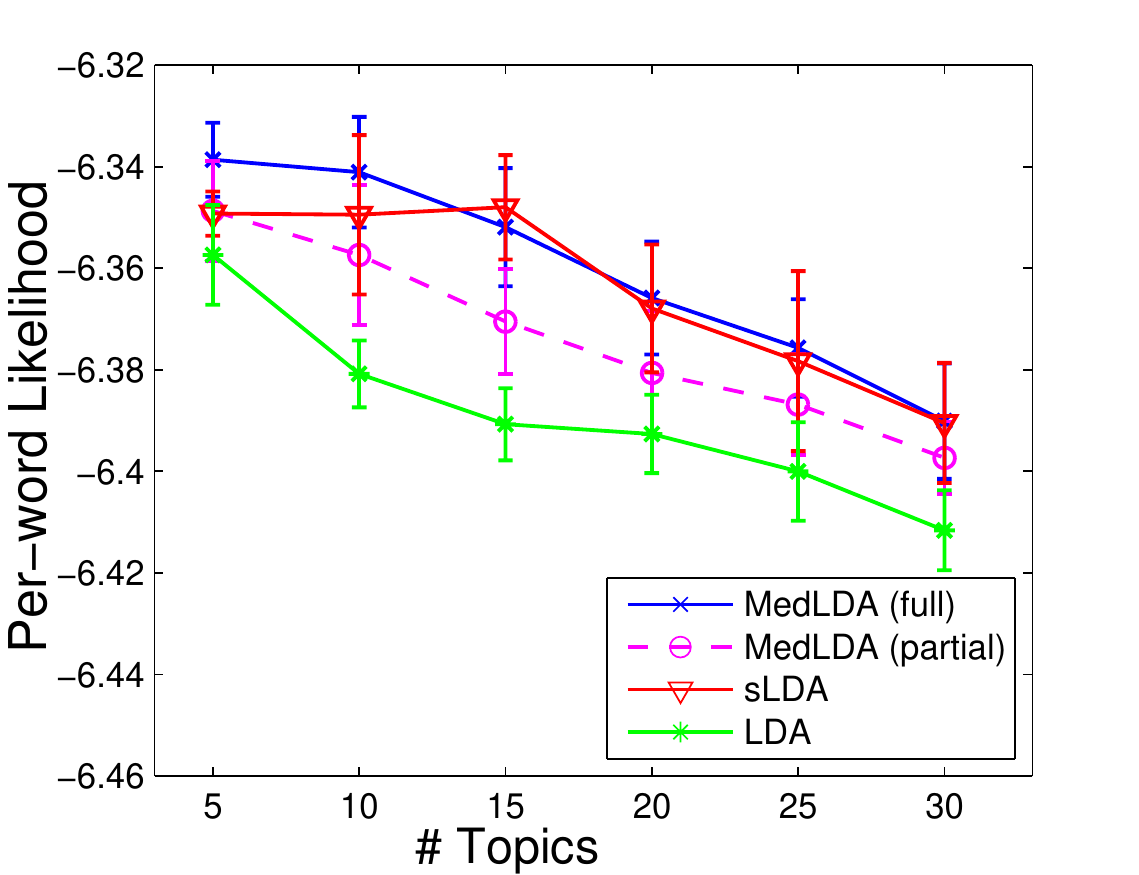}}}
\caption{\footnotesize Predictive R$^2$ (left) and
per-word likelihood (right) of different models on the movie review
dataset.} \vspace{-0.5cm}\label{fig_regression}
\end{figure}

Figure~\ref{fig_regression} shows the results together with the
per-word likelihood. We can see that the supervised MedLDA and sLDA
can get much better results than the unsupervised
LDA, which ignores supervised responses. 
By using max-margin learning, MedLDA (full) can get slightly better
results than the likelihood-based sLDA, especially when the number
of topics is small (e.g., $\leq 15$). Indeed, when the number of
topics is small, the latent representation of sLDA alone does not
result in a highly separable problem, thus the integration of
max-margin training helps in discovering a more discriminative
latent representation using the same number of topics. In fact, the
number of support vectors (i.e., documents that have at least one
non-zero lagrange multiplier) decreases dramatically at $T=15$ and
stays nearly the same for $T > 15$, which with reference to Eq.
(\ref{eq:phiMedLDAr}) explains why the relative improvement over
sLDA decreased as $T$ increases. This behavior suggests that MedLDA
can discover more predictive latent structures for {\it difficult},
non-separable problems.

For the two variants of MedLDA$^r$, we can see an obvious
improvement of MedLDA (full). This is because for MedLDA (partial),
the update rule of $\phi$ does not have the third and fourth terms
of Eq. (\ref{eq:phiMedLDAr}). Those terms make the max-margin
estimation and latent topic discovery attached more tightly.
Finally, a linear SVR on the empirical word frequency gets a pR$^2$
of 0.458, worse than those of sLDA and MedLDA.

\subsubsection{Time Efficiency}

For binary classification, MedLDA$^c$ is much more efficient than
sLDA, and is comparable with the LDA+SVM, as shown in Figure
\ref{fig:time}. The slowness of sLDA may be due to the mismatching
between its normal assumption and the non-Gaussian binary response
variables, which prolongs the E-step. 
For multi-class classification, the training time of MedLDA$^c$ is
mainly dependent on solving a multi-class SVM problem, and thus is
comparable to that of LDA. For regression, the training time of
MedLDA (full) is comparable to that of sLDA, while MedLDA (partial)
is more efficient.

\begin{figure}
\centering
\includegraphics[width =.6\columnwidth ]{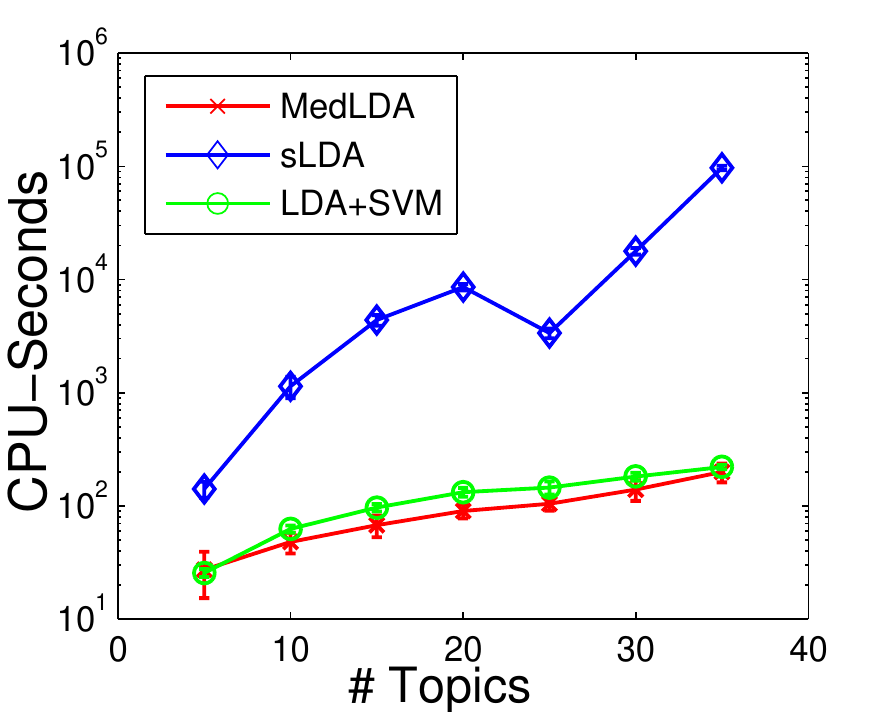}
\vspace{-0.5cm}\caption{Training time of different
models with respect to the number of topics for binary
classification.}\vspace{-0.5cm}\label{fig_time}
\end{figure}

\section{Related Work}

Latent Dirichlet Allocation (LDA) \citep{Blei:03} is a hierarchical
Bayesian model for discovering latent topics in a document
collection. LDA has found wide applications in information
retrieval, data mining, computer vision, and etc. The LDA is an
unsupervised model.

Supervised LDA \citep{Blei:07} was proposed for regression problem.
Although the sLDA was generalized to classification with a
generalized linear model (GLM), no results have been reported on the
classification performance of sLDA. One important issue that hinders
the sLDA to be effectively applied for classification is that it has
a normalization factor because sLDA defines a fully generative
model. The normalization factor makes the learning very difficult,
where variatioinal method or higher-order statistics must be applied
to deal with the normalizer, as shown in \citep{Blei:07}. Instead,
MedLDA applies the concept of margin and directly concentrates on
maximizing the margin. Thus, MedLDA does not need to define a fully
generative model, and the problem of MedLDA for classification can
be easily handled via solving a dual QP problem, in the same spirit
of SVM.

DiscLDA \citep{Simon:08} is another supervised LDA model, which was
specifically proposed for classification problem. DiscLDA also
defines a fully generative model, but instead of minimizing the
evidence, it minimizes the conditional likelihood, in the same
spirit of conditional random fields \citep{Lafferty:01}. Our MedLDA
significantly differs from the DiscLDA. The implementation of MedLDA
is extremely simple.

Other variants of topic models that leverage supervised information
have been developed in different application scenarios, including
the models for online reviews \citep{Titov:08,Branavan:08}, image
annotation \citep{He:08} and the credit attribution Labeled LDA
model \citep{Ramage:09}.

Maximum entropy discrimination (MED) \citep{Jaakkola:99} principe
provides an excellent combination of max-margin learning and
Bayesian-style estimation. Recent work \citep{Zhu:08b} extends the
MED framework to the structured learning setting and generalize to
incorporate structured hidden variables in a Markov network. MedLDA
is an application of the MED principle to learn a latent Dirichlet
allocation model. Unlike \citep{Westerdijk:00}, where a generative
model is degenerated to a deterministic version for classification,
our model is generative and thus can discover the latent topics over
document collections.

The basic principle of MedLDA can be generalized to the structured
prediction setting, in which multi-variant response variables are
predicted simultaneously and thus their mutual dependencies can be
explored to achieve globally consistent and optimal predictions. At
least two scenarios are within our horizon that can be directly
solved via MedLDA, i.e., the image annotation \citep{He:08}, where
neighboring annotation tends to be smooth, and the statistical
machine translation \citep{Zhao:06}, where tokens are naturally
aligned in word sentences.

\section{Conclusions and Discussions}

We have presented the maximum entropy discrimination LDA (MedLDA)
that uses the max-margin principle to train supervised topic models.
MedLDA integrates the max-margin principle into the latent topic
discovery process via optimizing one single objective function with
a set of {\it expected} margin constraints. This integration yields
a predictive topic representation that is more suitable for
regression or classification. We develop efficient variational
methods for MedLDA. The empirical results on movie review and 20
Newsgroups data sets show the promise of MedLDA on text modeling and
prediction accuracy.

MedLDA represents the first step towards integrating the max-margin
principle into supervised topic models, and under the general MedTM
framework presented in Section 3, several improvements and
extensions are in the horizon. Specifically, due to the nature of
MedTM's joint optimization formulation, advances in either
max-margin training or better variational bounds for inference can
be easily incorporated. For instance, the mean field variational
upper bound in MedLDA can be improved by using the tighter collapsed
variational bound \citep{Teh:nips06} that achieves results comparable
to collapsed Gibbs sampling \citep{Gibbs:04}. Moreover, as the
experimental results suggest, incorporation of a more expressive
underlying topic model enhances the overall performance. Therefore,
we plan to integrate and utilize other underlying topic models like
the fully generative sLDA model in the classification case. Finally, advanced in max-margin training would also results in more efficient training.


\section*{Acknowledgements}
This work was done while J.Z. was visiting CMU under a support from NSF DBI-0546594 and
DBI-0640543 awarded to E.X.; J.Z. is also supported by Chinese NSF
Grant 60621062 and 60605003; National Key Foundation R\&D Projects
2003CB317007, 2004CB318108 and 2007CB311003; and Basic Research
Foundation of Tsinghua National TNList Lab.

\appendix

\section*{Proof of Corollary 3}

In this section, we prove the corollary 3.

\begin{proof}
Since the variational parameters $(\gamma, \phi)$ are fixed when solving for $q(\eta)$, we can ignore the terms in $\mathcal{L}^{bs}$ that do not depend on $q(\eta)$ and get the function
\begin{eqnarray}
\mathcal{L}^{bs}_{\lbrack q(\eta) \rbrack } && \triangleq KL( q(\eta) \Vert p_0(\eta) ) - \sum_d E_{q} \lbrack \log p(y_d | \bar{Z}_d, \eta, \delta^2) \rbrack \nonumber \\
&& = KL( q(\eta) \Vert p_0(\eta) ) + \frac{1}{2\delta^2} \Big( E_{q(\eta)} \big\lbrack \eta^\top E \lbrack AA^\top \rbrack \eta - 2 \eta^\top \sum_{d=1}^D y_d  E \lbrack \bar{Z}_d \rbrack \big\rbrack \Big) + c, \nonumber
\end{eqnarray}
\noindent where $c$ is a constant that does not depend on $q(\eta)$.

Let $U(\xi, \xi^\star) = C \sum_{d=1}^D( \xi_d + \xi_d^\star)$.
Suppose $(q_0(\eta), \xi_0, \xi_0^\star)$ is the optimal solution of P1, then we have:
for any feasible $(q(\eta), \xi, \xi^\star)$,
$$\mathcal{L}^{bs}_{\lbrack q_0(\eta) \rbrack } + U(\xi_0, \xi_0^\star) \leq
\mathcal{L}^{bs}_{\lbrack q(\eta) \rbrack } + U(\xi, \xi^\star).$$

From Corollary~\ref{corollary:gaussian}, we conclude that the optimum
predictive parameter distribution is $q_0(\eta)=\mathcal{N}(\lambda_0, \Sigma)$, where
$\Sigma = (I + 1/\delta^2 E\lbrack A^\top A\rbrack)^{-1}$ does not depend on $q(\eta)$.
Since $q_0(\eta)$ is also normal, for any distribution\footnote{Although the feasible set of $q(\eta)$ in P1
is much richer than the set of normal distributions with the covariance matrix $\Sigma$, Corollary~\ref{corollary:gaussian}
shows that the solution is a restricted normal distribution. Thus, it suffices to consider only these normal
distributions in order to learn the mean of the optimum distribution.}
$q(\eta) = \mathcal{N}(\lambda, \Sigma)$, with several steps of algebra
it is easy to show that
$$\mathcal{L}^{bs}_{\lbrack q(\eta) \rbrack } = \frac{1}{2} \lambda^\top (I + \frac{1}{\delta^2} E\lbrack A^\top A\rbrack) \lambda - \lambda^\top (\sum_{d=1}^D \frac{y_d}{\delta^2} E\lbrack \bar{Z}_d \rbrack ) + c^\prime
= \frac{1}{2} \lambda^\top \Sigma^{-1} \lambda - \lambda^\top (\sum_{d=1}^D \frac{y_d}{\delta^2} E\lbrack \bar{Z}_d \rbrack ) + c^\prime, $$
where $c^\prime$ is another constant that does not depend on $\lambda$.

Thus, we can get: for any $(\lambda, \xi, \xi^\star)$, where
$$ (\lambda, \xi, \xi^\star) \in \lbrace (\lambda, \xi, \xi^\star):~y_d - \lambda^\top E\lbrack \bar{Z}_d \rbrack \leq
 \epsilon + \xi_d;~-y_d + \lambda^\top E\lbrack \bar{Z}_d \rbrack \leq
 \epsilon + \xi_d^\star;~\textrm{and}~\xi, \xi^\star \geq 0~\forall d \rbrace,$$
\noindent we have
$$\frac{1}{2} \lambda_0^\top \Sigma^{-1} \lambda_0  - \lambda_0^\top (\sum_{d=1}^D \frac{y_d}{\delta^2} E\lbrack \bar{Z}_d \rbrack ) + U(\xi_0, \xi_0^\star) \leq
\frac{1}{2} \lambda^\top \Sigma^{-1} \lambda - \lambda^\top (\sum_{d=1}^D \frac{y_d}{\delta^2} E\lbrack \bar{Z}_d \rbrack ) + U(\xi, \xi^\star),$$
\noindent which means the mean of the optimum posterior distribution under a Gaussian MedLDA is
achieved by solving a primal problem as stated in the Corollary.

\end{proof}

\bibliographystyle{plain}
\bibliography{MedLDA}
\end{document}